\documentclass[10pt,twocolumn,letterpaper]{article}

\usepackage{iccv}
\usepackage{times}
\usepackage{epsfig}
\usepackage{graphicx}
\usepackage{amsmath}
\usepackage{amssymb}

\usepackage{booktabs}
\usepackage{subcaption}

\usepackage{multirow} 

\usepackage{marvosym}

\usepackage[accsupp]{axessibility}

\usepackage[pagebackref=true,breaklinks=true,letterpaper=true,colorlinks,bookmarks=false]{hyperref}
\iccvfinalcopy

\def\iccvPaperID{2977} 

\ificcvfinal\fi

\begin{document}

\title{SparseNeRF: Distilling Depth Ranking for Few-shot Novel View Synthesis}

\author{Guangcong Wang \and
Zhaoxi Chen \and
Chen Change Loy \and
Ziwei Liu\textsuperscript{\Letter} \and
S-Lab, Nanyang Technological University \\
{\tt\small \{guangcong.wang,zhaoxi00,ccloy,ziwei.liu\}@ntu.edu.sg}
}

\ificcvfinal\thispagestyle{empty}\fi

\twocolumn[{
\renewcommand\twocolumn[1][]{#1}
\maketitle
\begin{center}
  \vspace{-0.2in}
  \centering
  \includegraphics[width=1.0\linewidth]{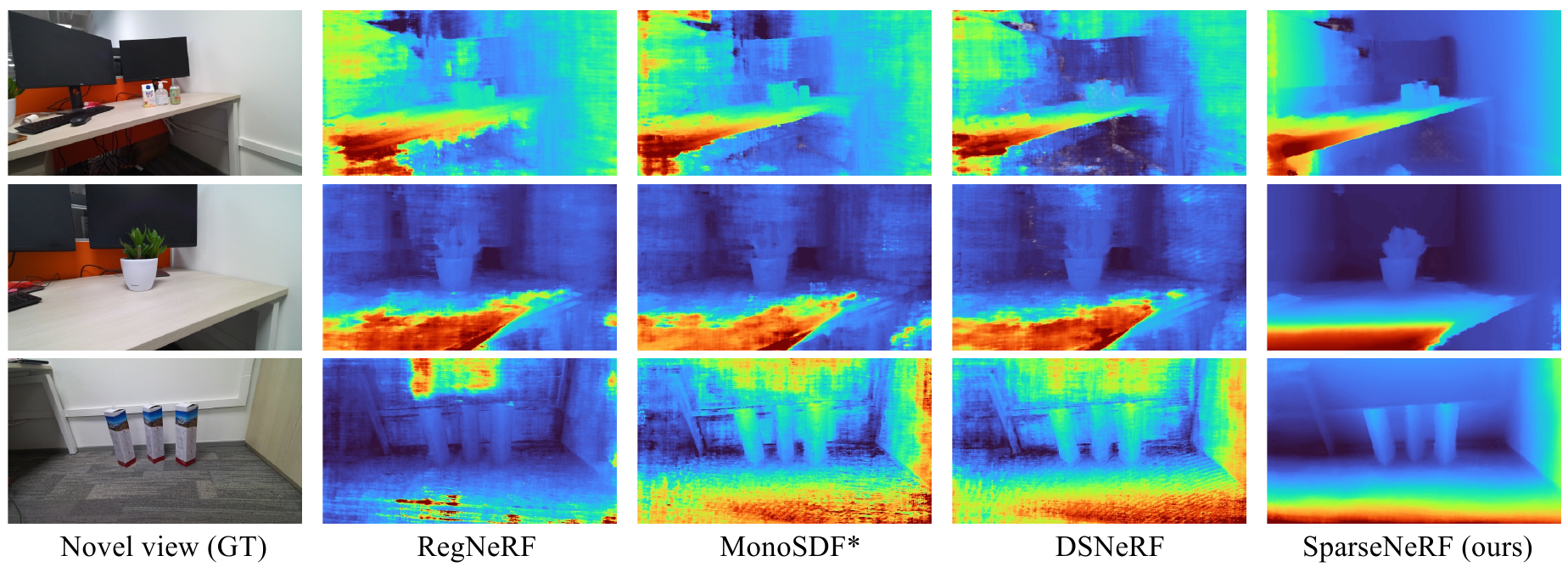}
    \captionof{figure}{Comparing the state-of-the-arts RegNeRF~\cite{niemeyer2022regnerf}, MonoSDF*~\cite{yu2022monosdf}, DSNeRF~\cite{deng2022depth}, and our SparseNeRF with three views for training (* denotes a re-implementation for our task). RegNeRF regularizes geometry with sparsity and continuity constraints. MonoSDF, and DSNeRF use scale-invariant depth constraints supervised by coarse depth maps. Our SparseNeRF uses robust depth ranking of coarse depth maps. SparseNeRF can synthesize realistic novel views and coherent geometric depth (Please refer to the supplementary material for rendered videos).} 
  \label{fig:demo1} 
\end{center}
}
]

\begin{abstract}
Neural Radiance Field (NeRF) significantly degrades when only a limited number of views are available. To complement the lack of 3D information, depth-based models, such as DSNeRF and MonoSDF, explicitly assume the availability of accurate depth maps of multiple views. They linearly scale the accurate depth maps as supervision to guide the predicted depth of few-shot NeRFs. However, accurate depth maps are difficult and expensive to capture due to wide-range depth distances in the wild. 

In this work, we present a new Sparse-view NeRF (\textbf{SparseNeRF}) framework that exploits depth priors from real-world inaccurate observations. The inaccurate depth observations are either from pre-trained depth models or coarse depth maps of consumer-level depth sensors. Since coarse depth maps are not strictly scaled to the ground-truth depth maps, we propose a simple yet effective constraint, a local depth ranking method, on NeRFs such that the expected depth ranking of the NeRF is consistent with that of the coarse depth maps in local patches. To preserve the spatial continuity of the estimated depth of NeRF, we further propose a spatial continuity constraint to encourage the consistency of the expected depth continuity of NeRF with coarse depth maps. Surprisingly, with simple depth ranking constraints, SparseNeRF outperforms all state-of-the-art few-shot NeRF methods (including depth-based models) on standard LLFF and DTU datasets. Moreover, we collect a new dataset NVS-RGBD that contains real-world depth maps from Azure Kinect, ZED 2, and iPhone 13 Pro. Extensive experiments on NVS-RGBD dataset also validate the superiority and generalizability of SparseNeRF. Code and dataset are available at \url{https://sparsenerf.github.io/}. 
\end{abstract}

\section{Introduction}
Neural radiance fields (NeRFs) \cite{mildenhall2020nerf,liu2020neural,barron2021mip,barron2022mip,sitzmann2021light} 
have made tremendous progress in generating photo-realistic novel views of scenes by optimizing implicit function representations given a set of 2D input views. However, in a wide range of real-world scenarios, collecting dense views of a scene is often expensive and time-consuming \cite{tancik2022block}. Therefore, it is necessary to develop few-shot NeRF methods that can be learned from sparse views without significant degradation in performance. 

Learning a NeRF from sparse views is a challenging problem due to under-constrained reconstruction conditions, especially in textureless areas. Directly applying NeRFs to few-shot scenarios suffers from dramatic degradation \cite{niemeyer2022regnerf}. Recently, some methods have greatly improved the performance of few-shot NeRF, which can be categorized into three groups. \textbf{1)} The first group \cite{niemeyer2022regnerf,kim2022infonerf,jain2021putting} is based on geometry constraints (sparsity and continuity regularizations) and high-level semantics. 
RegNeRF \cite{niemeyer2022regnerf} regularized the geometry and appearance of patches rendered from unobserved viewpoints, and annealed the ray sampling space. InfoNeRF \cite{kim2022infonerf} imposed an entropy constraint of the density in each ray and a spatial smoothness constraint into the estimated images. However, since a scene often contains multiple layouts (\textbf{Figure \ref{fig:demo1}}), sparsity and continuity geometric constraints of a few views cannot guarantee the complete 3D geometric reconstruction. \textbf{2)} The second group \cite{yu2021pixelnerf,chen2021mvsnerf} resorts to pre-training on similar scenes. For example, PixelNeRF \cite{yu2021pixelnerf} proposed to condition a NeRF on convolutional feature maps to learn high-level semantics from other scenes. \textbf{3)} The third group \cite{deng2022depth,yu2022monosdf,hu2023consistentnerf} exploits depth maps and makes a linearity assumption of depth maps to supervise few-shot NeRFs. For example, DSNeRF \cite{deng2022depth} exploited sparse 3D points generated by COLMAP \cite{schoenberger2016sfm} or accurate depth maps \cite{choi2016large} obtained by high-accuracy depth scanners and the Multi-View Stereo (MVS) algorithm. The depth maps are linearly scaled as supervision to guide the predicted depth of few-shot NeRFs. To use coarse depth maps, MonoSDF \cite{yu2022monosdf} uses a local patch-based scale-invariant depth constraint supervised by coarse depth maps instead of global depth maps. However, the scale-invariant depth constraint is strong for real-world coarse depth maps from pre-trained depth models or consumer-level depth sensors due to wide-range depth distances in the wild.

Along the third group, we wish to explore more robust 3D priors from coarse depth maps to complement the under-constrained few-shot NeRF.
To address this problem, we present SparseNeRF, a simple yet effective method that distills depth priors from pre-trained depth models \cite{ranftl2021vision} or inaccurate depth maps from consumer-level depth sensors (\textbf{Figure \ref{fig:coarse_depth}}), which can be easily obtained from real-world scenes.
Deriving useful depth cues from such pre-trained models is non-trivial. In particular, although single-view depth estimation methods have achieved good visual performance, thanks to large-scale monocular depth datasets and large ViT models, they cannot yield accurate 3D depth information due to coarse depth annotations, dataset bias, and ill-posed 2D single-view images. 
The inaccurate depth information contradicts the density prediction of a NeRF when reconstructing each pixel of a 3D scene based on volume rendering.
Directly scaling the coarse depth maps to a NeRF \cite{deng2022depth,yu2022monosdf} leads to inconsistent geometry against the expected depth of the NeRF. 

\begin{figure*}[t]
\vspace{-0.2in}
  \centering
  \includegraphics[width=1.0\linewidth]{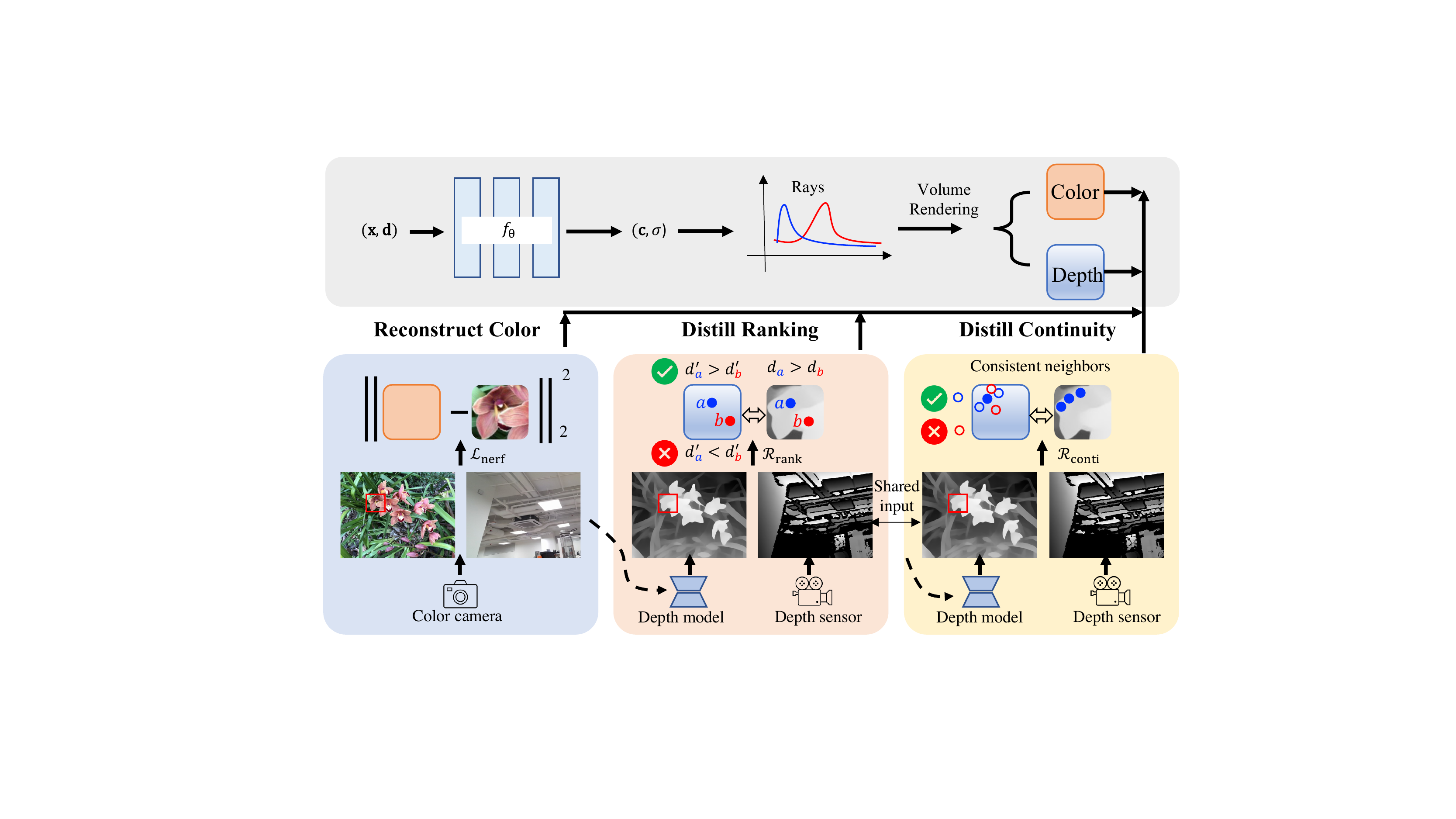}
  \vspace{-10pt}
  \caption{Framework Overview. SparseNeRF consists of two streams, i.e., NeRF and depth prior distillation. As for NeRF, we use Mip-NeRF as the backbone. we use a NeRF reconstruction loss $\mathcal{L}_{\mathrm{nerf}}$. As for depth prior distillation, we distill depth priors from a pre-trained depth model. Specifically, we propose a local depth ranking regularization and a spatial continuity regularization to distill robust depth priors from coarse depth maps.}
  \label{fig:framework}
  \vspace{-10pt}
\end{figure*}

Instead of directly supervising a NeRF with coarse depth priors, we relax hard depth constraints \cite{deng2022depth,yu2022monosdf} and distill robust local depth ranking from the coarse depth maps to a NeRF such that the depth ranking of a NeRF is consistent with that of coarse depth. That is, we supervise a NeRF with relative depth instead of absolute depth \cite{deng2022depth,yu2022monosdf}. To guarantee the spatial continuity of geometry, we further propose a spatial continuity constraint on depth maps such that the NeRF model imitates the spatial continuity of coarse depth maps. The accurate sparse geometry constraints from a limited number of views, combined with relaxed constraints including depth ranking regularization and continuity regularization, finally achieve promising novel view synthesis (\textbf{Figure \ref{fig:demo1}}).
It is noteworthy that SparseNeRF does not increase the running time during inference as it only exploits depth priors from pre-trained depth models or consumer-level sensors during the training stage (\textbf{Figure \ref{fig:framework})}.
In addition, SparseNeRF is a plug-and-play module that can be easily integrated into various NeRFs. 

The main contribution of this paper is \textbf{1)} SparseNeRF, a simple yet effective method that distills local depth ranking priors from pre-trained depth models. With the help of the local depth ranking constraint, SparseNeRF significantly improves the performance of few-shot novel view synthesis over the state-of-the-art models (including depth-based NeRF methods). To preserve the coherent geometry of a scene, we propose a spatial continuity distillation constraint that encourages the spatial continuity of NeRF to be similar to that of the pre-trained depth model. Both depth ranking prior and spatial continuity distillation are new in the literature on NeRF. 
\textbf{2)} Apart from SparseNeRF, we also contribute a new dataset, NVS-RGBD, which contains coarse depth maps from Azure Kinect, ZED 2, and iPhone 13 Pro.
\textbf{3)} Extensive experiments on the LLFF, DTU, and NVS-RGBD datasets demonstrate that SparseNeRF achieves a new state-of-the-art performance in few-shot novel view synthesis.

\section{Related Work}
\noindent\textbf{Neural Radiance Fields.} NeRF~\cite{mildenhall2020nerf,tao2023lidar,chen2023cunerf,peng2021pi} has made great success in synthesizing novel views of complex scenes due to good representation of neural networks \cite{he2016deep,wang2020grammatically,wang2018kalman,wang2019adaptively,dong2021efficientbert,li2022automated}. Block-NeRF \cite{tancik2022block}, CityNeRF  \cite{xiangli2021citynerf}, and Mega-NeRF \cite{turki2022mega} scaled the standard NeRF up to city-scale or urban-scale scenes. NeRF$--$ \cite{wang2021nerf}, GNeRF \cite{meng2021gnerf}, and BARF \cite{lin2021barf} relaxed the requirements of NeRFs and synthesized novel views without perfect camera poses. Some works attempted to improve NeRFs by considering anti-aliasing \cite{barron2021mip}, sparse 3D grids with spherical harmonics \cite{Fridovich-Keil_2022_CVPR}. Some methods  \cite{pumarola2021d,li2021neural,xian2021space} extended NeRFs to dynamic scenes. These methods do not focus on generating novel views with a few views. In this paper, we study sparse-view NeRF to reduce dense capture requirements in the real-world applications.

\noindent\textbf{Few-shot Novel View Synthesis.} There are increasing studies on few-shot novel view synthesis. Basically, the existing few-shot NeRF methods can be categorized into three groups. \textbf{First}, some methods exploit continuity constraints on geometry or object semantics. For example, 
RegNeRF \cite{niemeyer2022regnerf} imposed a continuity constraint on geometry and regularized the appearance of patches from unobserved viewpoints with a flow model.
InfoNeRF \cite{kim2022infonerf} proposed a ray entropy minimization regularization to encourage the density to be as sparse as possible along a ray, and used a ray information gain reduction regularization to constrain the continuous depth of neighbor rays.
\textbf{Second}, some methods attempt to pe-train a NeRF on other similar scenes and fine-tune the NeRF on the target scene. For example, PixelNeRF \cite{yu2021pixelnerf} conditioned a NeRF on image inputs in a fully convolutional manner, allowing the model to learn scene priors from other scenes and reduce the requirement of dense views.
MVSNeRF \cite{chen2021mvsnerf} leveraged plane-swept cost volumes for geometry-aware scene reasoning, and combined it with physically based volume rendering. Similar to PixelNeRF, MVSNeRF was first trained on other real scenes and was finetuned on target scenes to evaluate its effectiveness and generalizability. 
\textbf{Third}, depth-based models \cite{deng2022depth,yu2022monosdf} use available depth information to supervise the training of NeRFs. DSNeRF \cite{deng2022depth} exploited sparse 3D points generated by COLMAP \cite{schoenberger2016sfm} or accurate depth maps \cite{choi2016large} obtained by high-accuracy depth scanners. Different from these depth-based models, SparseNeRF distills robust depth ranking from pre-trained depth models or coarse depth maps from consumer-level sensors. As concurrent works, NeRDi \cite{deng2023nerdi} and NeuralLift-360 \cite{xu2023neurallift} also use ranking-based methods. However, they mainly focus on single-view setting while SparseNeRF focuses on sparse-view. SparseNeRF uses ranking loss to avoid inconsistent 3D geometry across different views while \cite{deng2023nerdi,xu2023neurallift} focuses on a soft geometric regularization on a single view (no cross-view consistency problem during training). Moreover, SparseNeRF introduces a new spatial continuity loss to distill spatial coherence from monocular depth estimators. In addition, some single-view synthesis methods \cite{xu2022sinnerf} either allow new generative objects from unseen views or focus on specific objects, e.g., face \cite {chan2022efficient} and human \cite{hu2023sherf}. Some NeRF-based methods \cite{hong2022eva3d,chen2023sd} extended to generation instead of reconstruction. Other AIGC models \cite{ho2020denoising,wang2022traditional,wang2022stylelight,zhuo2022fast,chen2022text2light,chen2022relighting, yang2023gp,li2023marconet,yang2022vtoonify} focus on the 2D generation or category-specific reconstruction \cite{chen2022structure}, which are different.

\section{Our Approach}
\begin{figure}[t]
\vspace{-0.2cm}
  \centering
  \includegraphics[width=1.0\linewidth]{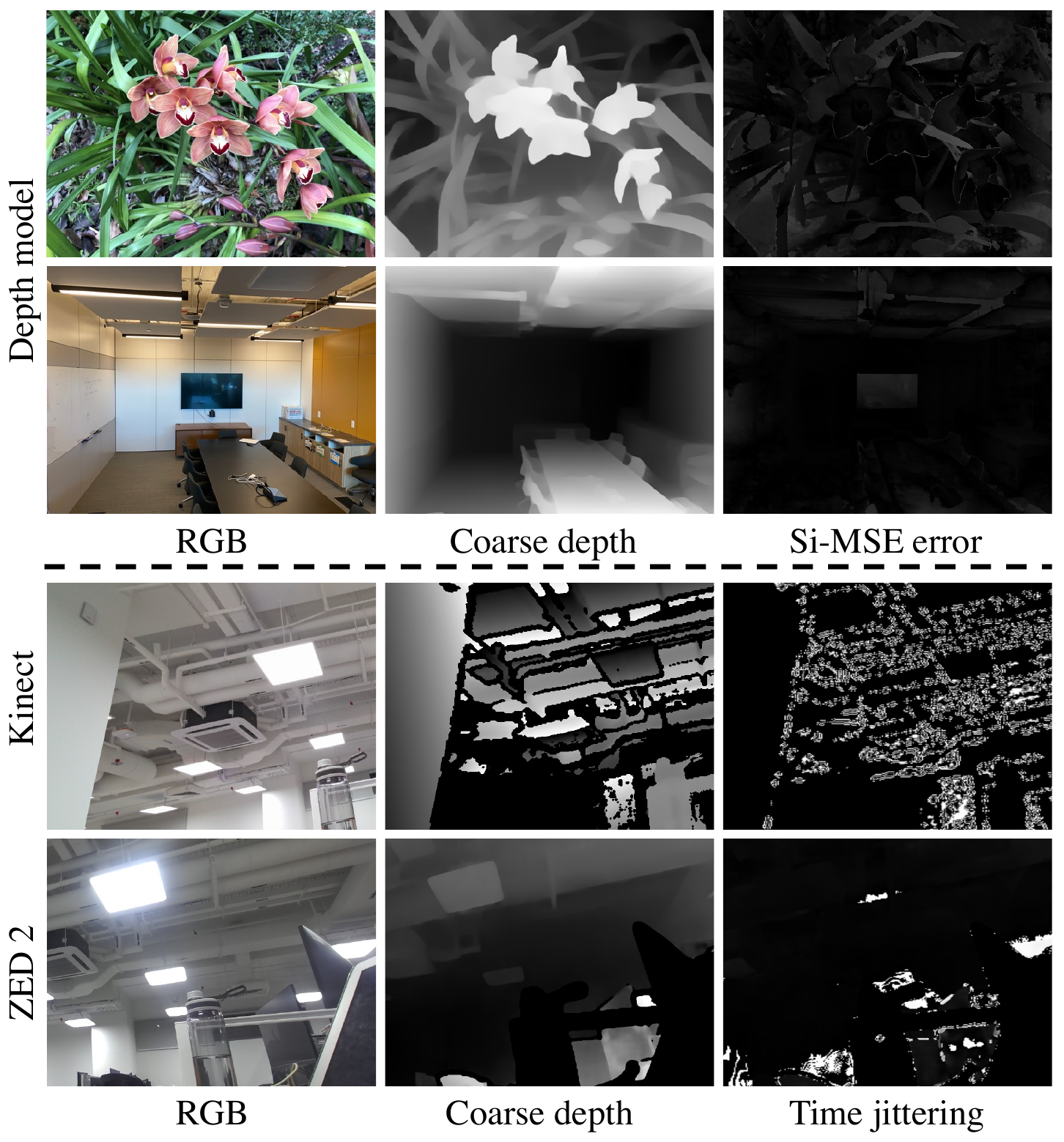}
  \vspace{-10pt}
  \caption{Two types of coarse depth maps. Top: predicted by a pre-trained depth model. Bottom: captured by  Azure Kinect and ZED 2 depth sensors. Si-MSE is a scale-invariant depth error. Time jittering refers to the difference between two depth maps captured by a static camera within a small time interval.}
  \label{fig:coarse_depth}
  \vspace{-0.1in}
\end{figure}

We present SparseNeRF to synthesize novel views given sparse view inputs. Single-view depth estimation is a long-standing computer vision task, aiming to predict a depth map given a single image. 
In this paper, we are interested in mining the depth priors encapsulated in pre-trained models of single-view depth estimation or coarse depth maps captured by consumer-level depth sensors.
However, due to coarse annotations of single-view depth maps, (\eg, user clicks \cite{chen2016single}, RGB-D \cite{silberman2012indoor,sturm2012benchmark}, and  laser/stereo \cite{menze2015object}), dataset bias, and imperfect depth estimation models, it is challenging to obtain accurate 3D depth estimation given 2D single-view images. As for consumer-level depth sensors, it still struggles to capture accurate depth maps (\textbf{Figure \ref{fig:coarse_depth}}). Driven by these observations, our goal is to make use of the coarse depth maps and distill useful depth priors to guide the learning of a NeRF.

\subsection{Preliminary and Problem Formulation}
\label{subsec:nerf}

\noindent\textbf{Neural Radiance Fields.} Let a NeRF \cite{mildenhall2020nerf} be a mapping function $f$ that maps a 3D spatial location $\mathbf{x}$ and a viewing direction $\mathbf{d}$ into a volume density $\sigma$ and a color value $\mathbf{c}$. The $f$ is a neural network consisting of eight perceptron (MLP) layers parameterized by $\theta$, which is given by $f_{\theta}:(\gamma(\mathbf{x}),\gamma(\mathbf{d}))\mapsto(\sigma, \mathbf{c})$, where $\gamma$ is a positional encoding. For each expected pixel color $\hat{C}(\mathbf{r})$, it is rendered by casting a ray $\mathbf{r}(t)=\mathbf{o}+t\mathbf{d}$ with near and far bounds $t_n$ and $t_f$. We evenly partition $[t_n, t_f]$ into $N$ points ($t_1, t_2, ..., t_N$) along a ray $\mathbf{r}$ and compute expected pixel color $\hat{C}(\mathbf{r})$ by $\hat{C}(\mathbf{r})=\sum_{i=1}^{N}\mathbf{c}_i^{*}$. The  weighted color $\mathbf{c}_i^{*}$ of a 3D point is computed by $\mathbf{c}_i^{*}=w_i\mathbf{c}_i$, where $\quad w_i=T_i(1-\mathrm{exp}(-\sigma_i\delta_i))$,  $T_i=\mathrm{exp}(-\sum_{j=1}^{i-1}\sigma_j\delta_j)$ and $\delta_i=t_{i}-t_{i-1}$. Therefore, the NeRF reconstruction loss can be formulated as 
\begin{equation}
\label{eq:reconstr}
\mathcal{L}_{\mathrm{nerf}}=\sum_{\mathbf{r}\in R}||\hat{C}(\mathbf{r})-C(\mathbf{r})||^{2},
\end{equation}
where $\hat{C}(\mathbf{r})$ are rendered color blended by $N$ samples. $C(\mathbf{r})$ is the ground-truth pixel color. We use coarse-to-fine sampling as discussed in the vanilla NeRF \cite{mildenhall2020nerf}. Here we omit the fine rendering for simplification. 

\noindent\textbf{Problem Formulation.} Vanilla NeRFs aim to learn a mapping function ${f_{\theta}}$ with $K_d$ dense views by optimizing a color reconstruction loss in \textbf{Eq. (\ref{eq:reconstr})}. In this work, we aim to study few-shot NeRF when only $K_s$ sparse views are available ($K_s \ll K_d$). Let $H$ and $W$ denote the height and width of an image, we have $HWK_s$ rays as color reconstruction constraints. Without considering view directions $\mathbf{d}$ and image continuity assumptions (i.e., $\mathbf{c}_i^{*}$ is independent of $\mathbf{c}_j^{*}$ when $i\neq j$), suppose we have a $N_c\times N_c\times N_c$ discrete cubic volume to be optimized, which contains $N_c^3$ weighted color variables $\mathbf{c}_i^{*}$. We have $N_c^3$ variables $\mathbf{c}_i^{*}$ and $HWK_s$ constraints. Ideally, we are able to solve a discrete cube of edge length $N_c=\sqrt[3]{HWK_s}$. That is, we can sample $\sqrt[3]{HWK_s}$ for each edge of a cubic volume. Take $H=W=512$ and $K_s=3$ as an example, we have $N_c\approx 92$, which is far from reconstructing a continuous radiance field. To address this under-constrained optimization problem of few-shot NeRF, an intuitive way is to introduce reasonable regularization terms to constrain few-shot NeRFs conditioned on input $\mathbf{x}$, $\mathbf{d}$, and $C(\mathbf{r})$. Considering these constraints, a general formulation is given by 
\begin{equation}
\label{eq:few-shot-nerf}
\mathcal{L}=\mathcal{L}_{\mathrm{nerf}}+\lambda \mathcal{R}(\mathbf{x},\mathbf{d},C(\mathbf{r})),
\end{equation}
where $\mathcal{R}$ is a regularization term.

\textbf{Remark.} RegNeRF \cite{niemeyer2022regnerf} and InfoNeRF \cite{kim2022infonerf} tackles this problem by introducing regularization terms on continuous depth constraint from unobserved viewpoints, sparsity of density on rays, and patch-based semantic constraint on color appearance. Depth-based models, such as DSNeRF \cite{deng2022depth} and MonoSDF \cite{yu2022monosdf}, directly regress depth by leveraging sparse 3D points or deriving ground-truth depth maps with accurate absolute depth maps. Different from them, we propose to use a robust relative depth regularization from coarse depth maps of pre-trained depth models or consumer-level depth sensors.

\subsection{Overview of SparseNeRF}
\label{subsec:overview}
The pipeline of SparseNeRF is illustrated in \textbf{Figure \ref{fig:framework}}. SparseNeRF mainly consists of four components, i.e., a neural radiance field (NeRF), a color reconstruction module, a depth ranking distillation module, and a spatial continuity distillation module. Specifically, we use Mip-NeRF \cite{barron2021mip} as the backbone and apply an MSE loss for color reconstruction $\mathcal{L}_{\mathrm{nerf}}$. As for depth prior distillation, we either use a pre-trained depth model to estimate depth maps or capture coarse depth maps with consumer-level depth sensors. We use a vision transformer (DPT) \cite{ranftl2021vision} that is trained on a large-scale mixed depth dataset (1.4M images with various depth annotations) that covers a wide range of scenes. Therefore, DPT is able to provide general depth priors. Since single-view depth estimation is coarse, we carefully design a local depth ranking regularization and a spatial continuity regularization, which distills robust depth priors from coarse maps to the NeRF.

\subsection{Local Depth Ranking Distillation}
\label{subsec:depth_ranking}
\begin{figure}[t]
\vspace{-0.2cm}
  \centering
  \includegraphics[width=1.0\linewidth]{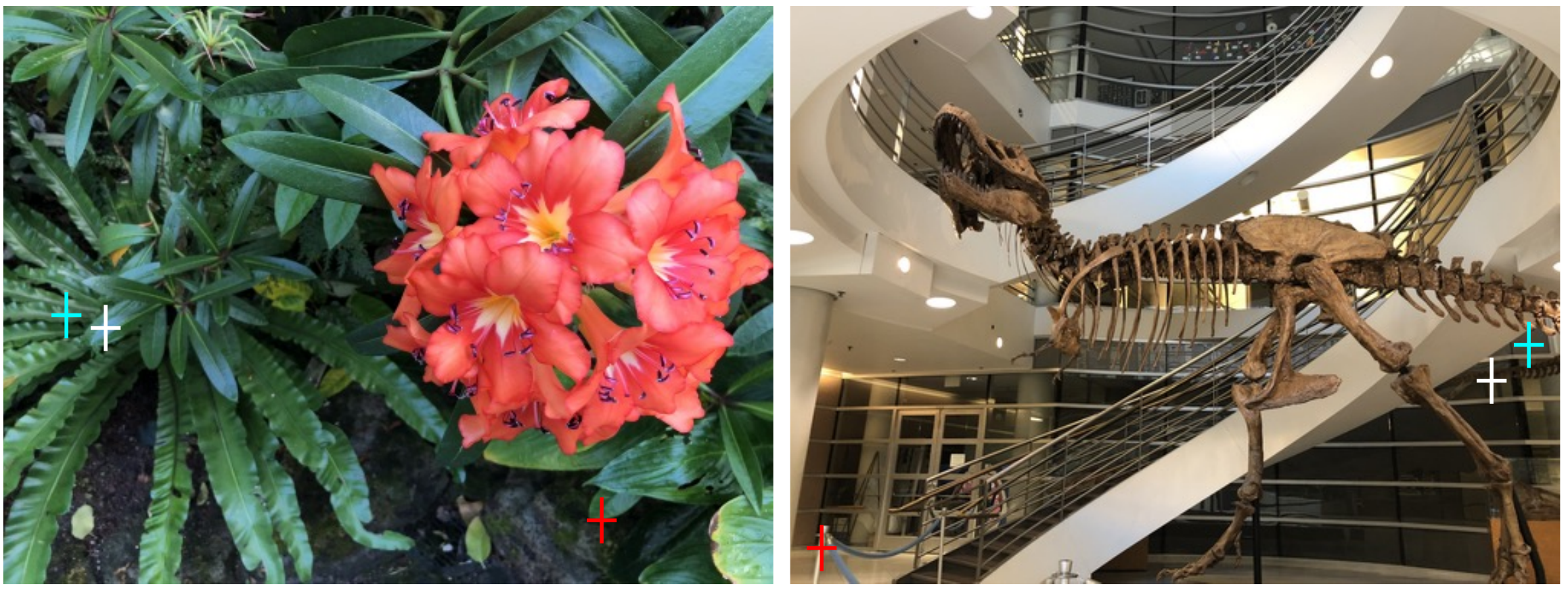}
  \vspace{-10pt}
  \caption{Motivation of local depth ranking. It is easy for depth estimation models to tell the white point is nearer than the cyan point, but it is hard to compare while and red points. Best viewed by zooming in.}
  \label{fig:why_local}
  \vspace{-10pt}
\end{figure}

Single-view depth estimation is a challenging computer vision task, which aims at predicting a depth map of a scene given a single image as input. Due to dataset bias, coarse depth annotations, and imperfect neural models, it is difficult to achieve accurate depth prediction. Depth maps captured by consumer-level sensors are also inaccurate due to wide-range depth distances. Directly using coarse depth maps to supervise a NeRF leads to ambiguous rendered novel view synthesis. To avoid the error of coarse depth maps, we relax the depth constraint and exploit the robust depth ranking prior. Given a pair of pixels in a single image, the depth ranking regularization only considers which point is nearer or farther. However, when the scene is complex, even depth ranking is not accurate. As shown in \textbf{Figure \ref{fig:why_local}}, it is easy for depth estimation models to compare the depth ranking of white and cyan points, but it is hard to estimate the depth ranking of white and red points. It implies that the depth ranking becomes unreliable as the spatial distance increases. 

Motivated by these observations, we propose a local depth ranking distillation method that distills depth ranking priors from coarse depth maps to a NeRF. On one hand, given a local patch $P$ of an RGB image $I$ with a pose $p$. We compute the depth $d_{\mathbf{r}}$ of the rays that trace from $P$ by $d_{\mathbf{r}}= \sum_{i=1}^{N}w_it_i$. On the other hand, we use pre-trained depth DPT \cite{ranftl2021vision} to estimate the depth of $I$ and crop a local patch $d_{\mathrm{dpt}}$ with the same spatial location as $d_{\mathbf{r}}$. We perform the depth ranking distillation by transferring the depth ranking knowledge from $d_{\mathrm{dpt}}$ to $d_{\mathbf{r}}$. Let $k1$ and $k2$ be the indices of 2D pixel coordinate of $d_{\mathrm{dpt}}$ and $d_{\mathbf{r}}$. The depth ranking regularization is given by
\begin{equation}
\label{eq:depth_ranking}
\mathcal{R}_{\mathrm{rank}}=\sum_{d_{\mathrm{dpt}}^{k1} \leq d_{\mathrm{dpt}}^{k2}}\mathrm{max}(d_{\mathbf{r}}^{k1}-d_{\mathbf{r}}^{k2}+m,0),
\end{equation}
where $m$ is a small margin that allows limited depth ranking errors. In \textbf{Eq. (\ref{eq:depth_ranking})}, we randomly sample two  depth pixels of $d_{\mathrm{dpt}}$, as denoted as $d_{\mathrm{dpt}}^{k1}$ and $d_{\mathrm{dpt}}^{k2}$, $d_{\mathrm{dpt}}^{k1} \leq d_{\mathrm{dpt}}^{k2}$. If the depth rankings of $d_{\mathrm{dpt}}$ and $d_{\mathbf{r}}$ are not consistent, i.e., $d_{\mathbf{r}}^{k1}>d_{\mathbf{r}}^{k2}$ and $d_{\mathrm{dpt}}^{k1} \leq d_{\mathrm{dpt}}^{k2}$, we punish the NeRF. We encourage the corresponding depth estimated by NeRF to satisfy the consistent depth ranking of the pre-trained model. We clip the large depth value and normalize the depth map for depth sensors. For depth models, we use relative inverse depth.

\subsection{Spatial Continuity Distillation}
\label{subsec:conti}
The local depth ranking constraint guarantees that the predicted depth map estimated by NeRF is consistent with the DPT's depth map. However, it does not constrain the spatial continuity of the depth map. We distill spatial continuity priors from the depth model DPT, which allows large displacement across several depth pixels. If neighbor depth pixels are continuous on the depth map of DPT, we constrain the corresponding depth pixels of NeRF to be continuous. The spatial continuity regularization is given by
\begin{equation}
\label{eq:depth_continuity}
\mathcal{R}_{\mathrm{conti}}=\sum_{k1}\sum_{d_{\mathrm{dpt}}^{k2} \in \mathrm{KNN}(d_{\mathrm{dpt}}^{k1})}\mathrm{max}(|d_{\mathbf{r}}^{k1}-d_{\mathbf{r}}^{k2}|-m^{'}, 0),
\end{equation}
where $\mathrm{KNN}(\cdot)$ returns k-nearest neighbors measured by depth values within a small region, \eg, $6\times6$ patch. $m^{'}$ is a small margin that allows small depth differences between neighbor pixels.

\noindent\textbf{Full Objective Loss.} The full objective loss of SparseNeRF is formulated by
\begin{equation}
\label{eq:full}
\mathcal{L}=\mathcal{L}_{\mathrm{nerf}}+\lambda\mathcal{R}_{\mathrm{rank}}+\gamma\mathcal{R}_{\mathrm{conti}},
\end{equation}
where $\lambda$ and $\gamma$ control the importance of the regularization terms. In practice, we set $\lambda=0.2$, $\gamma=0.02$, $m=m^{'}=1.0\times 10^{-4}$. 

\section{NVS-RGBD Dataset}
\label{sec:nvs-rgb}
The popular benchmark LLFF \cite{mildenhall2019local} on NeRFs is featured with forward-facing and object-centric. Moreover, we focus on consumer-level depth sensors instead of expensive scanners \cite{Choi2016}. We study stand-alone depth cameras instead of a complex calibrated structured multiple camera system \cite{dai2017scannet}, and collect from real-world scenes instead of synthesized 3D assets \cite{song2017semantic}. Few RGBD datasets satisfy all of the requirements. Therefore, we collect a new dataset NVS-RGBD that contains real-world depth maps captured by consumer-level depth sensors, including Azure Kinect, ZED 2, and iPhone 13 Pro. We collect 8 scenes for both Azure Kinect and ZED 2, and 4 scenes for iPhone 13 Pro. The depth maps of Azure Kinect often contain noises around object edges. The depth maps of ZED 2 look smooth but unstable and incorrect when observing time jittering (\textbf{Figure \ref{fig:coarse_depth}}). The poses are computed via COLMAP. We collect indoor scenes for ZED 2 and Kinect. For iPhone, we collect both indoor and outdoor scenes. All the scenes are object-centric. Depth maps obtained from sensors have different artifacts coming from the sensor noises. Please refer to the supplementary material for the details.

\section{Experiments}
\noindent\textbf{Datasets.} We conduct experiments on the LLFF \cite{mildenhall2019local}, DTU \cite{jensen2014large}, and NVS-RGBD datasets. LLFF contains 8 complex forward-facing scenes. Following RegNeRF \cite{niemeyer2022regnerf}, we use every 8-th image as the held-out test set, and evenly sample sparse views from the remaining images for training. Different from LLFF, DTU is an object-level dataset. Following PixNeRF \cite{yu2021pixelnerf,niemeyer2022regnerf}, we use the same 15 scenes in our experiments. On the DTU dataset, the background of these scenes is a white table or a black background, which has few textures. As mentioned in RegNeRF, we mask the background during inference to avoid background bias. We use the same evaluation protocol for a fair comparison. On the NVS-RGBD dataset, for each scene, we use 3 views for training and the rest views for inference. For LLFF and DTU, we use pre-trained depth maps to implement our method. For NVS-RGBD, we use coarse depth maps captured by depth sensors.

\noindent\textbf{Evaluation Metrics.} We adopt four evaluation metrics in the experiments, i.e., PSNR, SSIM \cite{wang2004image}, LPIPS \cite{zhang2018unreasonable}, and depth error \cite{deng2022depth}.  Depth error is a scale-invariant MSE derived by minimizing $||w\hat{d}+b-d||$, where $\hat{d}$ and $d$ denote coarse depth maps and pseudo-ground-truth depth maps predicted by dense-view NeRF. Note that we only use dense views to train a good NeRF to generate pseudo-ground-truth depth maps for evaluation.

\noindent\textbf{Implementation Details.} We implement SparseNeRF based on the official JAX \cite{jax2018github}. We use the Adam optimizer for the learning of SparseNeRF. We use an exponential learning rate, which decays from $2\times10^{-3}$ to $2\times 10^{-5}$. The batch size is set to 4096. For each scene, we use one GPU-v100 with 32G memory for both training and inference. We also implement the proposed SparseNeRF on GPU-v100 with 16G and RTX 2080 Ti with 11G. We reduce the batch size to 2048 and 1024 for these two types of GPUs, respectively. We use the same backbone and sampled points along rays compared with prior arts.

We train 90k iterations for each scene given three training views, which takes about 2 hours for each scene. For depth maps captured by Kinect and ZED 2, we mask uncertain regions (black regions in depth maps). More precisely, we sample near-far depth pairs in certain regions in local patches to optimize \textbf{Eqs. \ref{eq:depth_ranking}} and \textbf{\ref{eq:depth_continuity}} of the paper.

\begin{figure*}[t]
\vspace{-0.2in}
  \centering
  \includegraphics[width=1.0\linewidth]{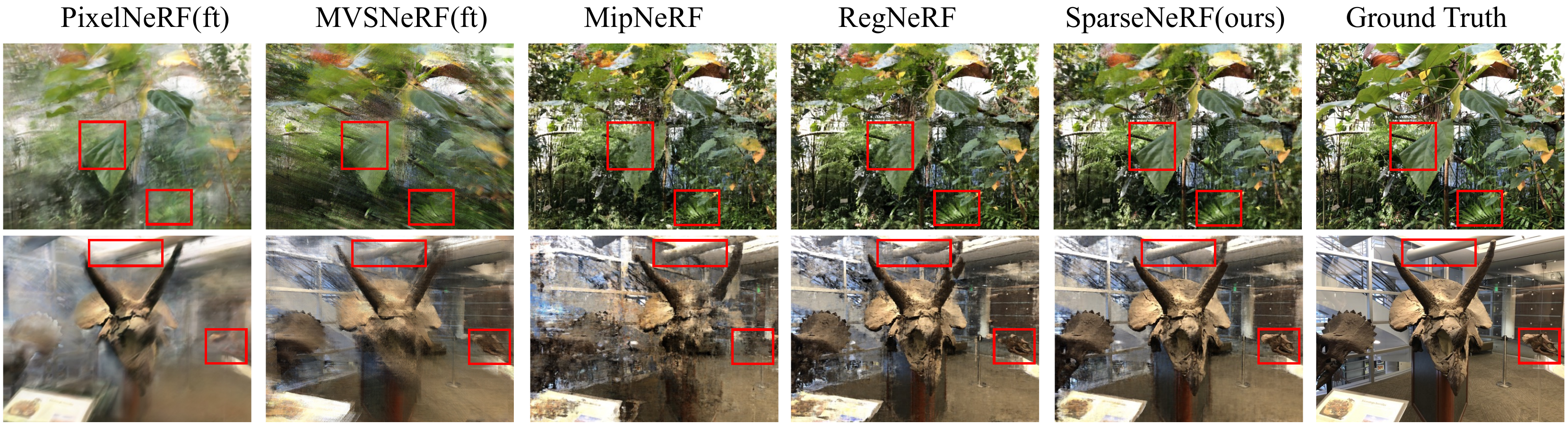}
  \vspace{-10pt}
  \caption{Visual comparisons on the LLFF dataset with three views. Red boxes denote compared regions. SparseNeRF achieves consistent improvement in different scenes.}
  \label{fig:llff_comparison}
\end{figure*}

\subsection{Comparisons on LLFF}
We compare SparseNeRF with state-of-the-art methods on the LLFF dataset, including SRF \cite{chibane2021stereo}, PixelNeRF \cite{yu2021pixelnerf}, MVSNeRF \cite{chen2021mvsnerf}, Mip-NeRF \cite{barron2021mip}, DietNeRF \cite{jain2021putting}, RegNeRF \cite{niemeyer2022regnerf}, DSNeRF \cite{deng2022depth} and MonoSDF\cite{yu2022monosdf}. Among them, SRF, PixelNeRF, and MVSNeRF are pre-trained on other similar scenes to exploit high-level semantics. Mip-NeRF is the state-of-the-art NeRF designed for dense-view training. InfoNeRF and RegNeRF mainly constrain sparsity and continuity of geometry or natural appearance semantics. More precisely, the compared methods can be classified into four groups. In the first group,  SRF, PixelNeRF, and MVSNeRF are pre-trained in the large-scale DTU datasets (88 scenes) and are directly tested on the LLFF dataset. In the second group, SRF, PixelNeRF, and MVSNeRF are pre-trained on DTU and fine-tuned on LLFF per scene. The third group includes Mip-NeRF, DietNeRF, and RegNeRF. They uses geometry continuity and semantic constraints. The fourth group distills knowledge from depth maps or sparse points of COLMAP (DSNeRF). The compared results are shown in \textbf{Table \ref{tab:Comparisons_llff}}. We can see that the proposed SparseNeRF achieves the best performance in PSNR, SSIM, and LPIPS. Note that the first and second groups have to pre-train on lots of other scenes. SparseNeRF achieves better results than DSNeRF and MonoSDF because the depth ranking is more robust than the absolute scale-invariant constraint. 

\begin{table}[t]
\caption{Quantitative Comparison on LLFF with three views. There are four groups. The first group denotes models that are pre-trained on other scenes and tested on a target scene (P). The second group requires further finetuning on a target scene (P\&FT). The third group uses geometry and semantic constraints (geo. \& sem.). The fourth group distills depth knowledge (depth KD). }\label{tab:Comparisons_llff}
\vspace{-15pt}
\begin{center}
\scalebox{0.75}{
\begin{tabular}{c|c|c|c|c}
\toprule
~&\textbf{Setting}&\textbf{PSNR $\uparrow$}&\textbf{SSIM $\uparrow$}&\textbf{LPIPS $\downarrow$}\\
\hline
\hline
SRF \cite{chibane2021stereo}& 
& 12.34    
&0.250 
&0.591 	\\

PixelNeRF \cite{yu2021pixelnerf}&P
& 7.93      
&0.272  
&0.682 \\
MVSNeRF \cite{chen2021mvsnerf}&
& 17.25       
&0.557  
&0.356\\
\hline
\hline
SRF ft \cite{chibane2021stereo}&
& 17.07 
&0.436 
&0.529 \\

PixelNeRF ft \cite{yu2021pixelnerf}&P\&FT
& 16.17    
&0.438  
&0.512\\
MVSNeRF ft \cite{chen2021mvsnerf}&
& 17.88       
&0.584   
&0.327 \\
\hline
\hline
Mip-NeRF \cite{barron2021mip}&
& 14.62   
&0.351  
&0.495 \\

DietNeRF \cite{jain2021putting}& geo. \& sem.
& 14.94     
&0.370   
&0.496 \\
RegNeRF \cite{niemeyer2022regnerf}&
& 19.08      
&0.587   
&0.336 \\
\hline
\hline
MonoSDF*\cite{yu2022monosdf}&\multirow{3}{*}{depth KD}
& 18.45     
&0.565   
&0.388 \\
DSNeRF\cite{deng2022depth}&
& 18.94     
&0.582   
&0.362 \\
\textbf{SparseNeRF (Ours)}& 
& \textbf{19.86} 
& \textbf{0.624} 
&\textbf{0.328}	\\
\bottomrule
\end{tabular}}
\end{center}
\vspace{-0.3in}
\end{table}

We provide qualitative analysis on LLFF, as shown in \textbf{Figure \ref{fig:llff_comparison}}. It is observed that PixelNeRF and MVSNeRF tend to generate ambiguous pixels. The reason is that they integrate CNN features into NeRFs. CNN features can provide high-level semantics to infer under-constrained textures but suffer pixel-level inference. MipNeRF is designed for dense-view synthesis without considering the under-constrained problem, leading to degraded geometry artifacts. Compared with RegNeRF, SparseNeRF uses robust depth priors from  pre-trained depth models, which can handle challenging scene geometry with better performance.

\begin{figure*}[t]
\vspace{-0.1in}
  \centering
  \includegraphics[width=1.0\linewidth]{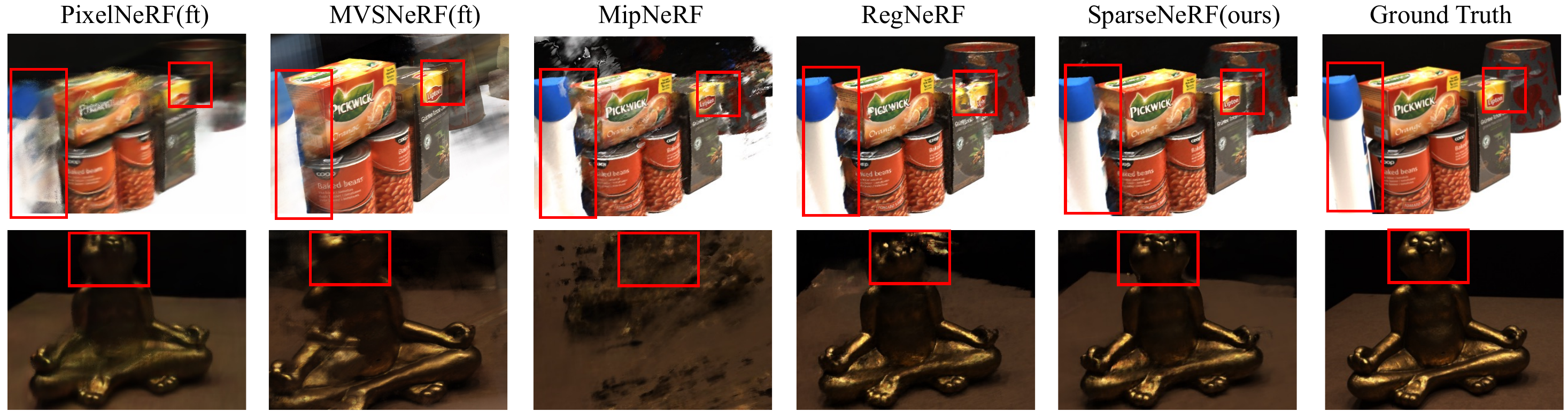}
  \vspace{-10pt}
  \caption{Visual comparisons on the DTU dataset with three views. Red boxes denote compared regions. SparseNeRF achieves consistent improvement in different scenes.}
  \label{fig:dtu_comparison}
  \vspace{-10pt}
\end{figure*}

\begin{figure*}[t]
  \centering
  \includegraphics[width=1.0\linewidth]{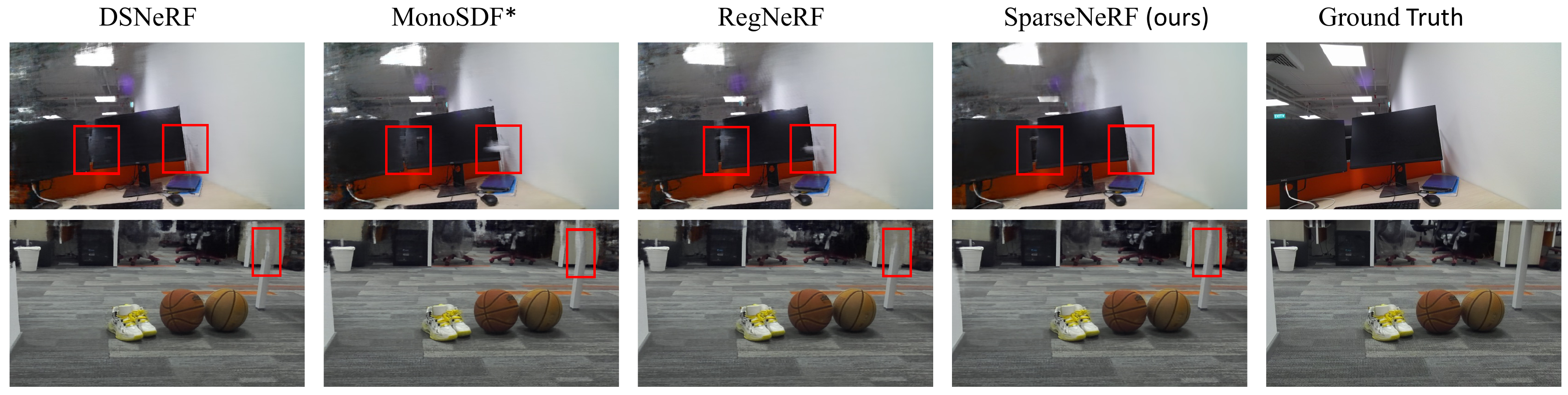}
  \vspace{-10pt}
  \caption{Visual comparisons on the NVS-RGBD dataset with three views. Red boxes denote compared regions. SparseNeRF achieves consistent improvement in different scenes.}
  \label{fig:NVS_comparison}
  \vspace{-10pt}
\end{figure*}

\subsection{Comparisons on DTU}

\begin{table}[h]
\caption{Quantitative Comparison on DTU with three views. The first group is pre-trained on other scenes and tested on a target scene (P). The second group requires further finetuning on a target scene (P\&FT). The third group uses geometry and semantic constraints (geo. \& sem.). The fourth group distills depth knowledge (depth KD).}\label{tab:Comparisons_dtu}
\vspace{-15pt}
\begin{center}
\scalebox{0.78}{
\begin{tabular}{c|c|c|c|c}
\toprule
~&\textbf{Setting}&\textbf{PSNR $\uparrow$}&\textbf{SSIM $\uparrow$}&\textbf{LPIPS $\downarrow$}\\
\hline
\hline
SRF \cite{chibane2021stereo}& 
& 15.32   
&0.671 	 
&0.304 \\
PixelNeRF \cite{yu2021pixelnerf}&P
& 16.82      
&0.695   
&0.270  \\
MVSNeRF \cite{chen2021mvsnerf}&
& 18.63      
&0.769 
&0.197 \\
\hline
\hline
SRF ft \cite{chibane2021stereo}&
& 15.68    
&0.698 	 
&0.281 \\
PixelNeRF ft \cite{yu2021pixelnerf}&P\&FT
& 18.95      
&0.710  
&0.269 \\
MVSNeRF ft \cite{chen2021mvsnerf}&
& 18.54       
&0.769   
&0.197 \\
\hline
\hline
Mip-NeRF \cite{barron2021mip}&
& 8.68    
&0.571 	 
&0.353 \\
DietNeRF \cite{jain2021putting}& geo. \& sema.
& 11.85    
&0.633	  
&0.314 \\
RegNeRF \cite{niemeyer2022regnerf}&
& 18.89       
&0.745    
&\textbf{0.190} \\
\hline
\hline
MonoSDF* \cite{yu2022monosdf}& \multirow{3}{*}{depth KD}
& 18.92    
&0.748	  
&0.237 \\
DSNeRF \cite{deng2022depth}& 
& 16.90    
&0.570	  
&0.450 \\
\textbf{SparseNeRF}\textbf{(Ours)}&
& \textbf{19.55}      
&\textbf{0.769}    
&0.201\\
\bottomrule
\end{tabular}}
\end{center}
\end{table}
\begin{figure}[h!]
  \centering
  \includegraphics[width=1.0\linewidth]{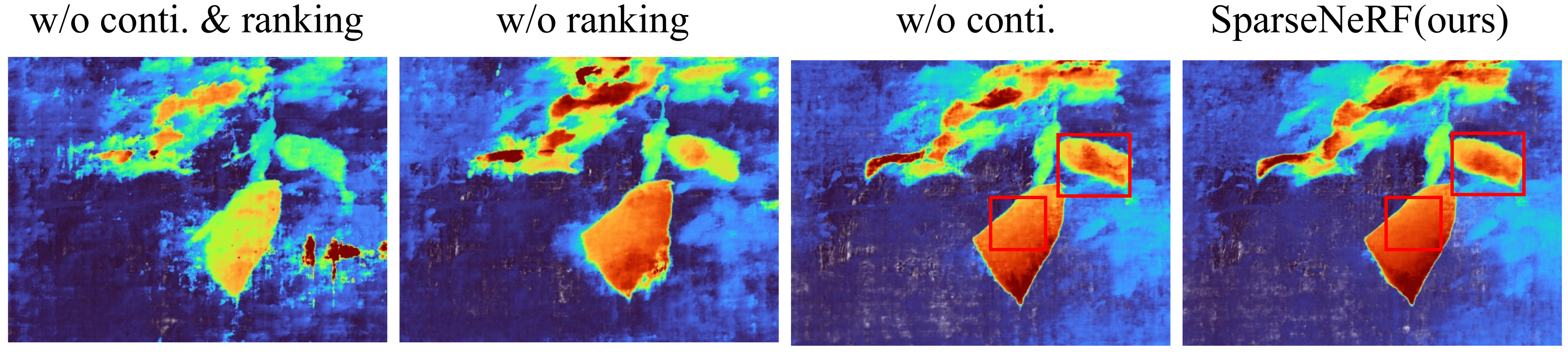}
  \caption{Ablation study on visual effect. The three columns denote ``w/o continuity and ranking distillation", ``w/o continuity distillation", and ``w/ continuity and ranking distillation" (SparseNeRF, ours), respectively.}
  \label{fig:ablation_study}
\end{figure}
Similar to the LLFF dataset, we conduct four-group comparisons on the DTU dataset, as shown in \textbf{Table \ref{tab:Comparisons_dtu}}. In this first group, given a target scene, SRF, PixelNeRF, and MVSNeRF are pre-trained on other DTU scenes and are tested on the target scene. In the second group, these three methods are further fine-tuned on the target scene. In the third group, we compare Mip-NeRF, DietNeRF, and RegNeRF. Note that for the pre-trained models, they split DTU as a training set (88 scenes) and a test set (15 scenes). Since the training set and the test set contains similar scenes, a test scene shares a similar distribution (e.g., similar backgrounds) with training scenes. Pre-training on other scenes greatly benefits a target scene. Therefore, SRF, PixelNeRF, and MVSNeRF also greatly achieve promising results. Finally, we compare depth-based DSNeRF \cite{deng2022depth} and MonoSDF \cite{yu2022monosdf}. DSNeRF studies two kinds of depth information, i.e., sparse 3D points generated by COLMAP \cite{schoenberger2016sfm} and ground-truth depth maps obtained by high-accuracy depth scanners and the Multi-View Stereo (MVS) algorithm. The depth maps of NVS-RGBD are captured by consumer-level sensors and contain lots of noise, which is hard to be applied to DSNeRF. Therefore, we adopt sparse 3D points for DSNeRF. Compared with these methods, SparseNeRF still achieves the best performance without pre-training on other scenes. We provide a qualitative analysis, as shown in \textbf{Figure \ref{fig:dtu_comparison}}. SparseNeRF achieves better visual results than previous state-of-the-art methods.

\subsection{Comparisons on NVS-RGBD}
\label{subsec:nvs-rgbd}
We then implement three state-of-the-art methods on our new NVS-RGBD dataset, including RegNeRF \cite{niemeyer2022regnerf}, DSNeRF \cite{deng2022depth}, and MonoSDF \cite{yu2022monosdf}. All of them are optimized per scene without pre-training on the other scenes. RegNeRF adopted continuous geometry assumptions from unobserved viewpoints. DSNeRF uses a global scale-invariant depth constraint to supervise NeRFs. MonoSDF mainly focuses on implicit surface reconstruction. It is assumed that local patches of depth maps from monocular cameras satisfy $d^{'}=wd+b$ where $d$ is predicted by the pre-trained model and $d^{'}$ is then computed by the NeRF. In \textbf{Table \ref{tab:nvs_rgbd}}, we observe that SparseNeRF achieves better performance compared with these prior arts. We provide a qualitative analysis in \textbf{Figure \ref{fig:NVS_comparison}}, demonstrating the superiority of SparseNeRF over previous state-of-the-art methods.

\begin{table}[]
\caption{Comparison with State of the Arts on NVS-RGBD. * denotes a new implementation from related tasks.}\label{tab:nvs_rgbd}
\vspace{-20pt}
\begin{center}
\scalebox{0.8}{
\begin{tabular}{c|c|c|c|c|c}
\toprule
                        & Metrics          & PSNR$\uparrow$ & SSIM$\uparrow$ & LPIPS$\downarrow$ &depth err$\downarrow$\\ \hline
\multirow{4}{*}{\rotatebox[origin=c]{90}{Kinect}} & RegNeRF \cite{niemeyer2022regnerf}          &  25.78    &  0.840    & 0.242  &  5.9$\times 10^{-3}$  \\ \cline{2-6} 
                        & DSNeRF \cite{deng2022depth}          &  25.91    &  0.838    &  0.238   & 4.8 $\times 10^{-3}$ \\ \cline{2-6} 
                        & MonoSDF* \cite{yu2022monosdf}        & 25.50     & 0.838     &   0.239  & 5.5$\times 10^{-3}$ \\ \cline{2-6} 
                        & \textbf{SparseNeRF(ours)} &  \textbf{26.28}    &  \textbf{0.850}    &   \textbf{0.232} & \textbf{3.9}$\mathbf{\times 10^{-3}}$  \\ \hline \hline
\multirow{4}{*}{\rotatebox[origin=c]{90}{ZED 2}}  & RegNeRF \cite{niemeyer2022regnerf}         &  24.98    &  0.781    &  0.271  & 3.5 $\times 10^{-3}$ \\ \cline{2-6} 
                        & DSNeRF \cite{deng2022depth}          & 24.97     & 0.789     &  \textbf{0.262}   & 3.1$\times 10^{-3}$ \\ \cline{2-6} 
                        & MonoSDF* \cite{yu2022monosdf}        & 24.54     & 0.784     &  0.275   & 3.2 $\times 10^{-3}$\\ \cline{2-6} 
                        & \textbf{SparseNeRF(ours)} & \textbf{26.22}     & \textbf{0.804}     &  \textbf{0.262}  & \textbf{2.4} $\mathbf{\times 10^{-3}}$  \\ \bottomrule
\end{tabular}}
\end{center}
\vspace{-15pt}
\end{table}

\begin{table}[h]
\caption{Comparison of Depth Ranking and Depth scaling.}\label{tab:ranking_vs_scaling}
\vspace{-15pt}
\begin{center}
\scalebox{0.87}{
\begin{tabular}{c|c|c|c|c}
\toprule
LLFF&PSNR$\uparrow$&SSIM$\uparrow$&LPIPS$\downarrow$&depth err$\downarrow$\\
\hline
baseline &19.08&0.587&0.336&10.1$\times 10^{-3}$\\
\hline
w/ depth scaling&18.45&0.565&0.388&8.9$\times 10^{-3}$\\
\textbf{w/ depth ranking}&\textbf{19.86}&\textbf{0.624}&\textbf{0.328}&\textbf{6.3}$\mathbf{\times 10^{-3}}$\\
\bottomrule
\end{tabular}}
\end{center}
\vspace{-20pt}
\end{table}

\subsection{Ablation Studies and Further Analyses}
\noindent\textbf{Effectiveness of Depth Distillation.} To validate the effectiveness of the depth distillation, we isolate the depth constraint and keep other modules unchanged. As shown in the first and third rows of \textbf{Table \ref{tab:ranking_vs_scaling}}, it is observed that without the local depth ranking distillation, SparseNeRF degrades on PSNR, SSIM, LPIPS, and depth error, which demonstrates the effectiveness of the depth distillation. We further study the depth maps from unobserved views. The ablation study of visual effect is shown in \textbf{Figure \ref{fig:ablation_study}}. It shows that depth ranking greatly improves performance in geometry. Spatial continuity regularization further improves the detailed coherence of scenes.

\noindent\textbf{Further Analyses on Depth Ranking and Depth Scaling.} To validate the robustness of local depth ranking, we compare depth ranking regularization and depth scaling regularization on LLFF. We use the idea from MonoSDF \cite{yu2022monosdf}, which mainly focuses on implicit surface reconstruction. With the specific design for surface shapes, MonoSDF slightly degrades the performance of RGB view synthesis. We re-implement it as depth scaling. Since the single-view depth estimation is not accurate, the linear depth scaling constraint is too strong such that depth distillation misleads the learning of the NeRF. \textbf{Figure \ref{fig:coarse_depth}} shows two cases of scale-invariant errors. Instead, depth ranking relaxes the constraint and only focuses on depth comparison, which is more robust than depth scaling. As shown in \textbf{Table \ref{tab:ranking_vs_scaling}}, depth ranking performs better than depth scaling on LLFF. 

\noindent\textbf{Impact of Different Pre-trained Depth Models.} We test our method on three depth estimation models, i.e., MiDaS small, DPT Hybrid, and DPT Large. As shown in \textbf{Table \ref{tab:diff_depth_models}}, all of the depth models are better than the baseline. DPT Hybrid and DPT Large achieve comparable results, which are better than MiDaS small. 

\begin{table}[t]
\caption{Impact of different pre-trained depth models.}\label{tab:diff_depth_models}
\vspace{-15pt}
\begin{center}
\scalebox{0.94}{
\begin{tabular}{c|c|c|c|c}
\toprule
LLFF&PSNR$\uparrow$&SSIM$\uparrow$&LPIPS$\downarrow$&depth err$\downarrow$\\
\hline
Baseline&  19.08 &0.587&0.336& 10.1$\times10^{-3}$\\
\hline
MiDaS small& 19.35 & 0.592 &0.333&6.9$\times10^{-3}$\\
DPT Hybrid& 19.86 & 0.625 & 0.326&6.3$\times10^{-3}$\\
DPT Large& 19.86 & 0.624& 0.328&6.3$\times10^{-3}$\\
\bottomrule
\end{tabular}}
\end{center}
\vspace{-20pt}
\end{table}

\noindent\textbf{FreeNeRF \cite{Yang2023FreeNeRF} potentially complements SparseNeRF.} FreeNeRF and SparseNeRF improve few-shot NeRF in different ways. FreeNeRF introduces a frequency regularization while SparseNeRF distills depth priors from pre-trained general depth estimators. We implement SparseNeRF on FreeNeRF and the results significantly improve, as shown in \textbf{Table \ref{tab:FreeNeRF}}.
\begin{table}[h]
\caption{FreeNeRF potentially complement SparseNeRF.}\label{tab:FreeNeRF}
\vspace{-10pt}
\begin{center}
\scalebox{0.85}{
\begin{tabular}{c|c|c|c}
\toprule
LLFF&PSNR$\uparrow$&SSIM$\uparrow$&LPIPS$\downarrow$\\
\hline
FreeNeRF &19.60&0.614&0.302\\
\textbf{FreeNeRF+our depth distillation}&\textbf{20.26}&\textbf{0.646}&\textbf{0.292}\\
\bottomrule
\end{tabular}}
\end{center}
\vspace{-20pt}
\end{table}

\begin{figure}[h]
  \centering
  \includegraphics[width=1.0\linewidth]{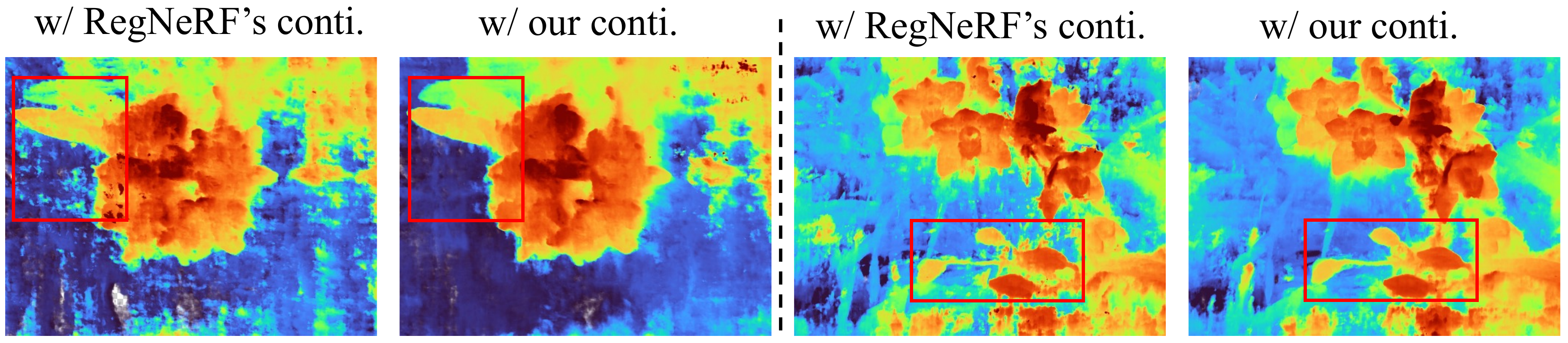}
  \caption{Visual comparison on continuity regulation.}
  \label{fig:ablation_study_conti}
\end{figure}

\noindent\textbf{User Study.} To show that the proposed spatial continuity regularization improves the 3D consistency in 3D space, we conducted a user study. For each scene, we rendered videos for both with and without spatial continuity regularization terms. Users are asked to select the better-quality video according to 3D consistency/coherence and completeness (good geometry, few floaters). We evaluate 8 scenes on LLFF and there are 27 participants. The result shows 88.9\% of users prefer ``w/ conti" and 11.1\% prefer ``w/o conti". To evaluate the advantages of SparseNeRF on 3D consistency and shape completeness, we conduct another user study. Users are asked to evaluate ``baseline+RegNeRF's continuity term" and ``baseline+our continuity term". The 27 users participated in the study and 8 scenes on LLFF are evaluated. The result shows 87.5\% of users prefer ``baseline+our continuity term" and 12.5\% prefer ``baseline+RegNeRF's continuity term". RegNeRF encourages the overall patch to be as smooth as possible without considering the sharp edges of objects, which conflicts with the reconstruction from sparse views. Our continuity regulation distills the neighbor relationship from depth models is more accurate. We show two examples of depth maps in \textbf{Figure \ref{fig:ablation_study_conti}}.

\noindent\textbf{Discussion.} SparseNeRF could have a wide range of applications due to its promising results. We provide two applications on iPhone 13 Pro. Please refer to the project page for more experiments. Like UNISURF \cite{oechsle2021unisurf,yu2022monosdf}, we also evaluate the SparseNeRF using Chamfer distance, showing the competitive results with MonoSDF and UNISURF, especially for complex scenes.

\section{Conclusion}
In this paper, we propose a SparseNeRF framework that synthesizes novel views with sparse view inputs. To tackle the under-constrained few-shot NeRF problem, we propose a local depth ranking regularization that distills the depth ranking prior from coarse depth maps. To preserve the spatial continuity of geometry, we propose a spatial continuity regularization that distills the depth continuity priors. With the proposed depth priors, SparseNeRF achieves a new state-of-the-art performance on three datasets.

\noindent\textbf{Limitation.} SparseNeRF significantly improves the performance of few-shot NeRF, but cannot be generalized to occluded views that are unobserved in the training views.

\noindent\textbf{Potential Negative Impact.} SparseNeRF focuses on scene synthesis. It might be misused to create misleading content.

\noindent\textbf{Acknowledgement.} This study is supported under the RIE2020 Industry Alignment Fund Industry Collaboration Projects (IAF-ICP) Funding Initiative, as well as cash and in-kind contribution from the industry partner(s). It is also supported by Singapore MOE AcRF Tier 2 (MOE-T2EP20221-0012, MOE-T2EP20221-0011) and NTU NAP Grant.

{\small
\bibliographystyle{ieee_fullname}
\bibliography{egbib}
}
\clearpage
\appendix

\twocolumn[{
\renewcommand\twocolumn[1][]{#1}
\maketitle
\begin{center}
  \vspace{-20pt}
  \centering
  \includegraphics[width=1.0\linewidth]{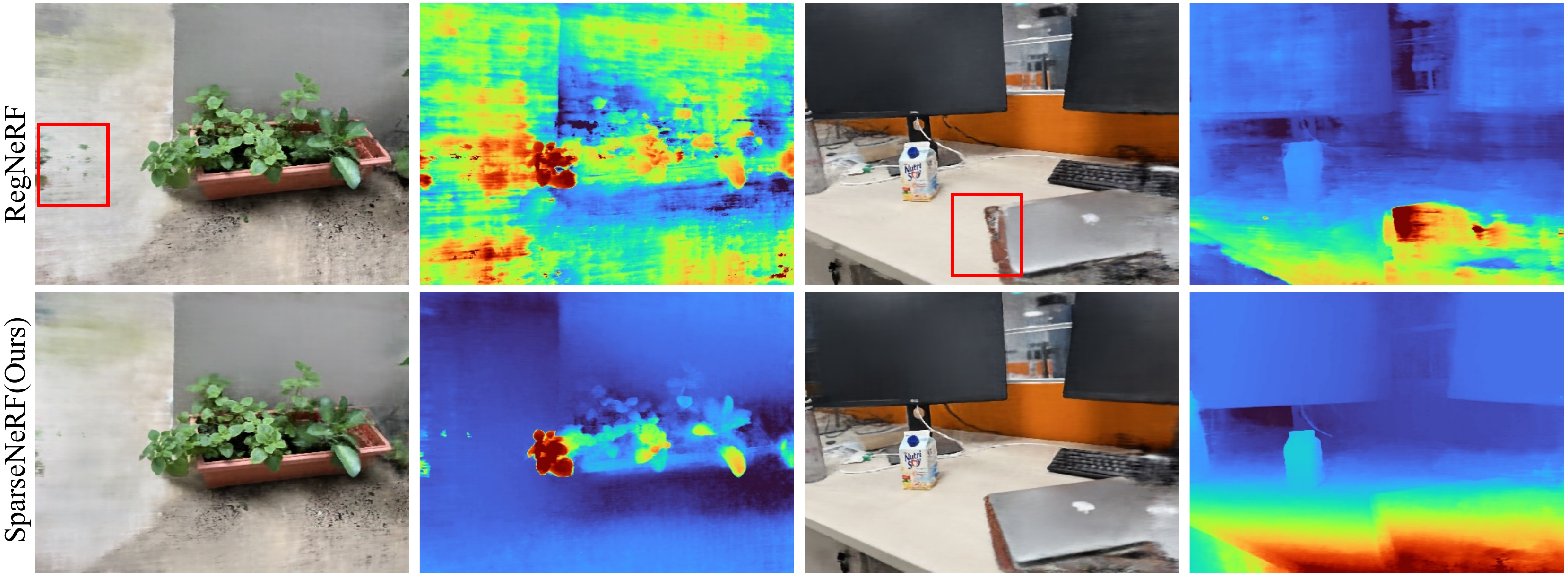}
  \vspace{-10pt}
    \captionof{figure}{Applications on iPhone 13 Pro. We use three RGDB images for training. Depth maps are captured by iPhone's LiDAR. Comparing the state-of-the-art RegNeRF~\cite{niemeyer2022regnerf} and the proposed SparseNeRF with three views for training, our method can extract useful depth priors from the model and synthesize realistic novel views and coherent geometric depth even though only three training views are available (Please refer to the demo videos for rendered videos).} 
  \label{fig:iphone_app} 
\end{center}
}
]

\begin{figure*}[h]
  \centering
  \includegraphics[width=\linewidth]{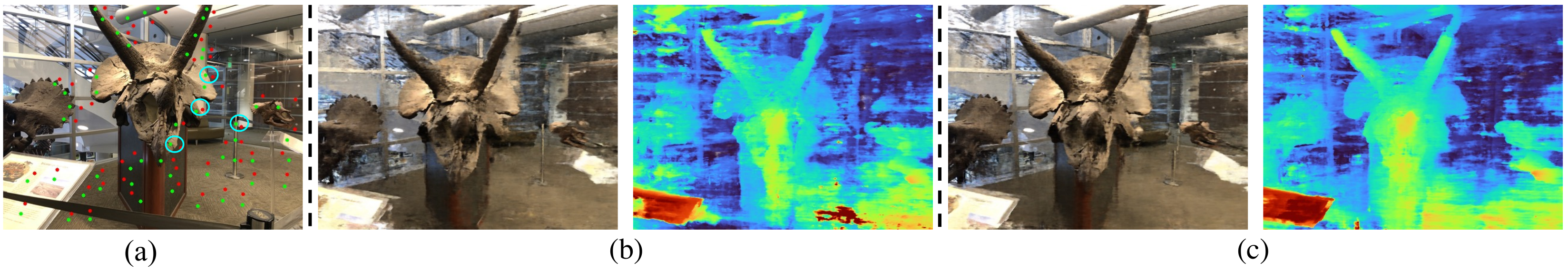}
  \caption{Application with simple user-click depth annotations. Annotators are required to click near-far depth pairs in a local region (e.g., near-far pairs in cyan circles). Green points are nearer while red points are farther. (a) User-click depth annotations. (b) Without user-click depth annotations. (c) With user-click depth annotations. User-click depth annotations provide sparse depth rankings, which tailors for the proposed SparseNeRF.}
  \label{fig:refinement}
\end{figure*}

\section*{A. Applications}
\subsection*{A1. Few-shot NeRF on iPhone} 

Apple's latest products (iPhone 12 Pro and higher versions, and iPad Pro now feature built-in LiDAR sensors that can capture depth maps. We use the portrait mode of iPhone 13 Pro to collect RGBD datasets. We train RegNeRF and SparseNeRF with three views (RGB images and depth maps are aligned), and rendered novel views, as shown in \textbf{Figure \ref{fig:iphone_app}}. RegNeRF does not recover a correct geometry while the proposed SparseNeRF significantly improves the reconstruction of geometry. The main reason is that RegNeRF adds a geometric continuity constraint from unseen viewpoints such that neighbor depth pixels should be as close as possible. This regularization does not fully exploit the depth priors of objects/scenes. Instead, the proposed SparseNeRF distills the depth ranking and the spatial continuity of depth maps captured by an iPhone's LiDAR, which achieves promising rendered results.

\subsection*{A2. Few-shot NeRF with Simple User-click Depth Annotations} The SparseNeRF significantly improves few-shot novel view synthesis with the help of coarse depth maps, obtained by pre-trained depth models and consumer-level sensors. What if the depth rankings of partial regions of depth maps are incorrect in some cases? Do we have any ideas to fix this issue? We provide an application that fixes incorrect depth rankings of pre-trained depth or sensors when we do not trust these coarse depth maps. As shown in \textbf{Figure \ref{fig:refinement} (a)}, users can easily click near-far depth pairs in a local region, e.g., depth pairs in cyan circles. Green points are nearer while red points are farther. In \textbf{Figure \ref{fig:refinement} (b)} and \textbf{Figure \ref{fig:refinement} (c)}, we train a few-shot NeRF without and with user-click depth annotations, respectively. It is observed that the performance of the few-shot NeRF significantly improves. In practice, 30$\sim$40 depth pairs of each view lead to significant improvement. In real-world scenes, one may only need to annotate partial regions for refinement. User-click depth annotations provide sparse depth rankings, which tailors for the proposed SparseNeRF. In real-world applications, occlusion often occurs in complex scenes. Even with dense views for training, some regions are observed from a limited number of views. Therefore, this application helps dense-view NeRF to refine novel view synthesis in these cases.

\section*{B. Detailed Descriptions of the new Dataset NVS-RGBD}
In \textbf{Section \ref{sec:nvs-rgb}}, we give a brief introduction to our new dataset NVS-RGBD. In this material, we provide a detailed description. Different from previous methods that derive accurate and expensive ground-truth depth maps by pre-trained NeRF trained with dense views \cite{fan2022unified} or captured by high-accuracy depth scanners \cite{deng2022depth}, our goal is to develop a good NeRF with cheap or even free coarse depth maps, i.e., pre-trained single-view depth estimation models and consumer-level depth sensors. To this end, we use Microsoft Azure Kinect, ZED 2, and iPhone 13 Pro to collect a new dataset NVS-RGBD. We collect 4 scenes for iPhone 13, and 8 scenes for ZED 2 and Kinect, respectively. The dataset is similar to the LLFF dataset that focuses on forward-facing scenes. For each scene, we evenly sample 3 views for training and the held-out views are for testing.

\begin{figure*}[h]
  \centering
  \includegraphics[width=\linewidth]{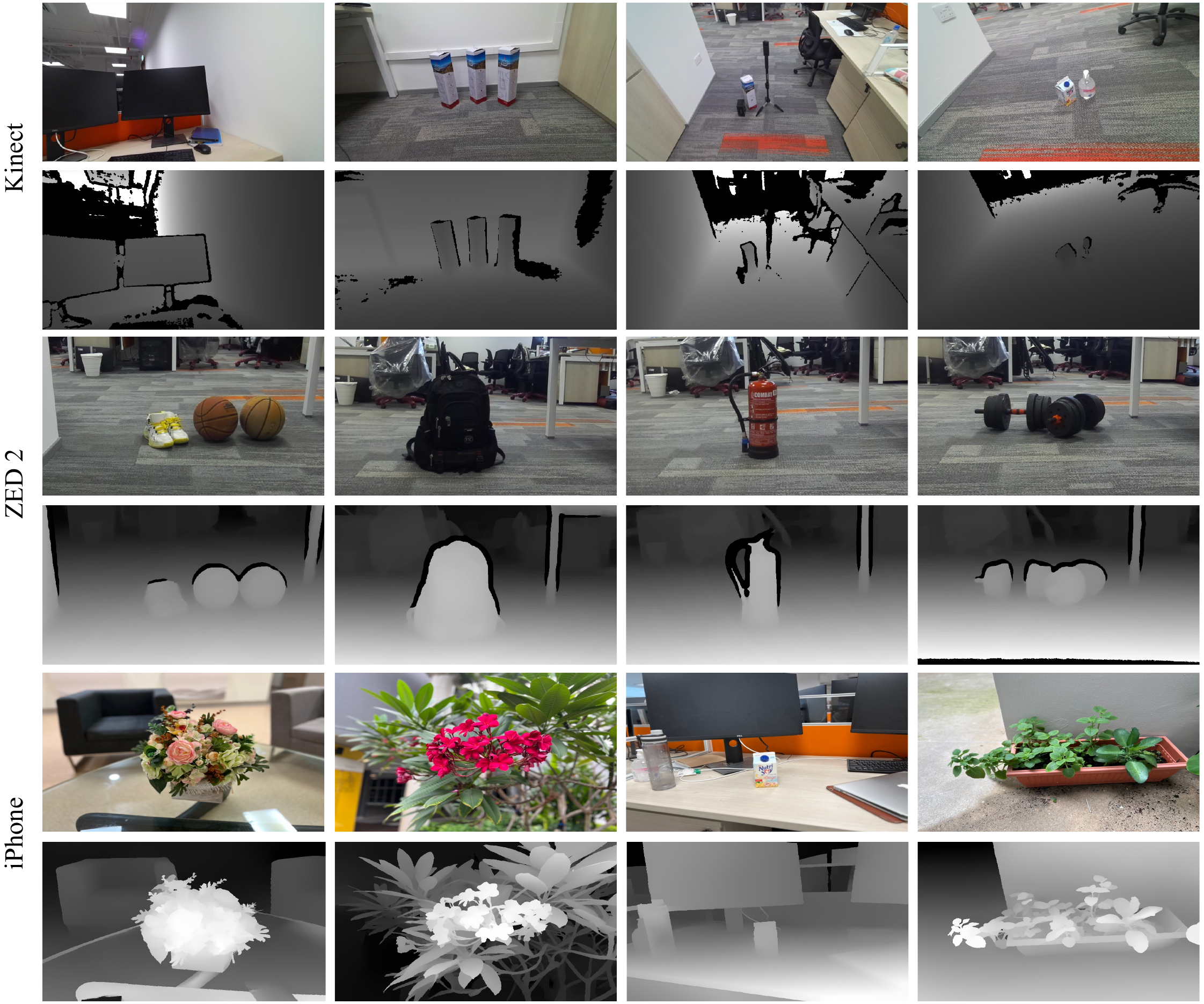}
  \caption{Examples of the NVS-RGBD dataset. It contains RGBD images captured by Azure Kinect, ZED 2, and iPhone.}
  \label{fig:nvs_examples}
\end{figure*}

\begin{figure*}[t]
  \centering
  \includegraphics[width=1.0\linewidth]{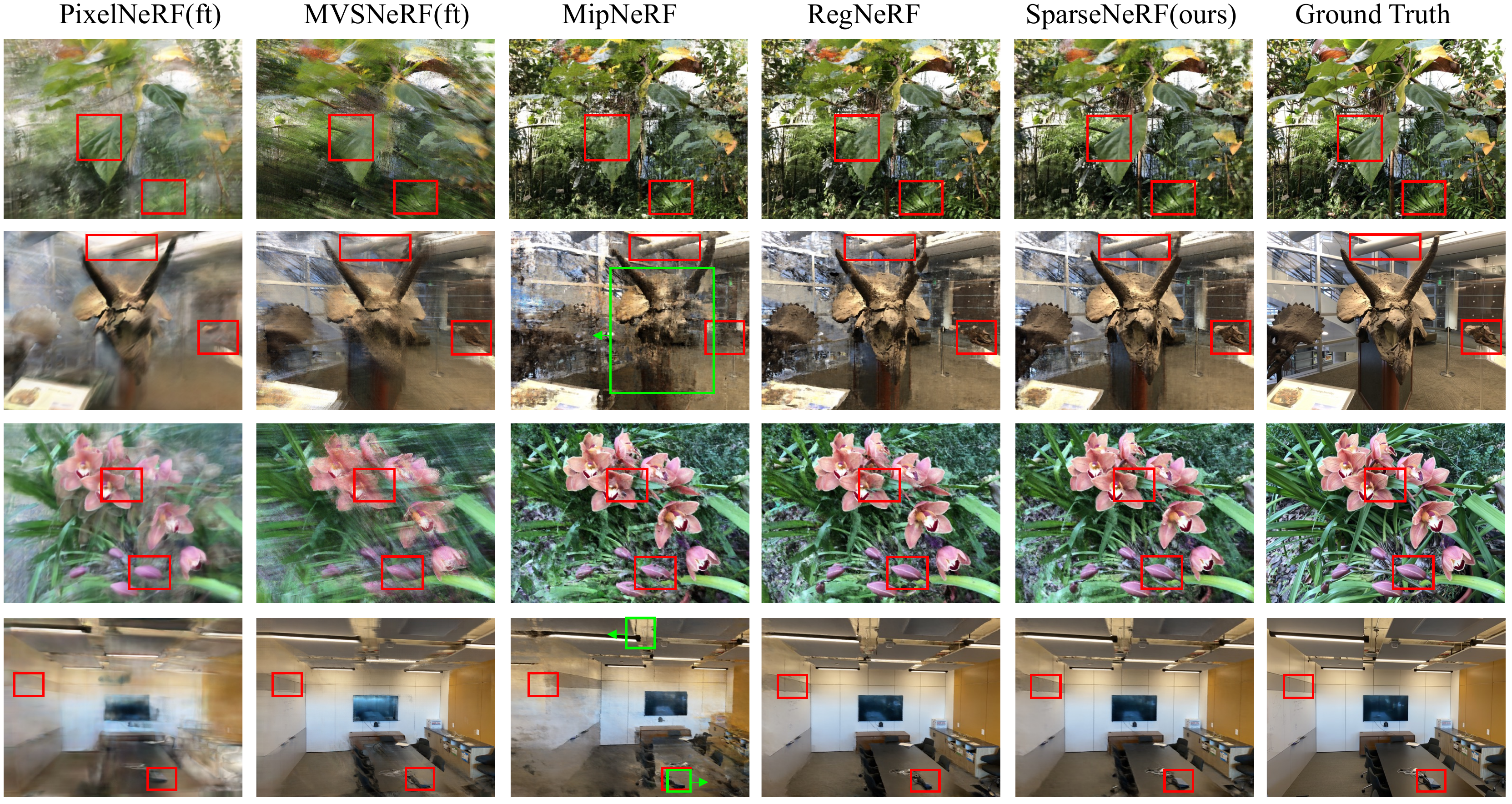}
  \vspace{-10pt}
  \caption{Visual comparisons on the LLFF dataset with three views. Red boxes denote compared regions. Green boxes denote shifted objects. SparseNeRF achieves consistent improvement in different scenes.}
  \label{fig:llff_comparison_supp}
\end{figure*}

\begin{figure*}[t]
  \centering
  \includegraphics[width=1.0\linewidth]{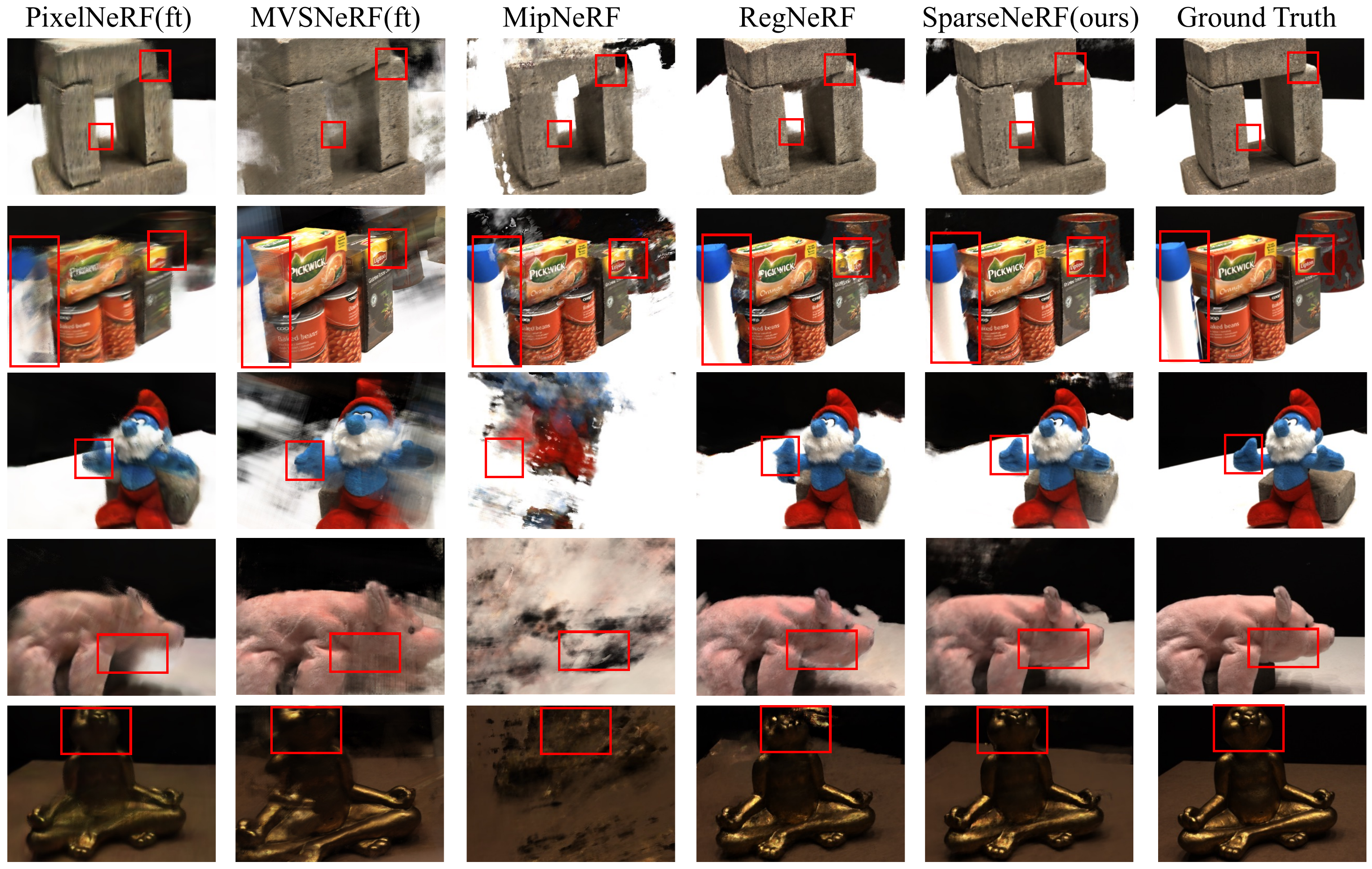}
  \vspace{-10pt}
  \caption{Visual comparisons on the DTU dataset with three views. Red boxes denote compared regions. SparseNeRF achieves consistent improvement in different scenes.}
  \label{fig:dtu_comparison_supp}
  \vspace{-10pt}
\end{figure*}

\begin{figure*}[t]
  \centering
  \includegraphics[width=1.0\linewidth]{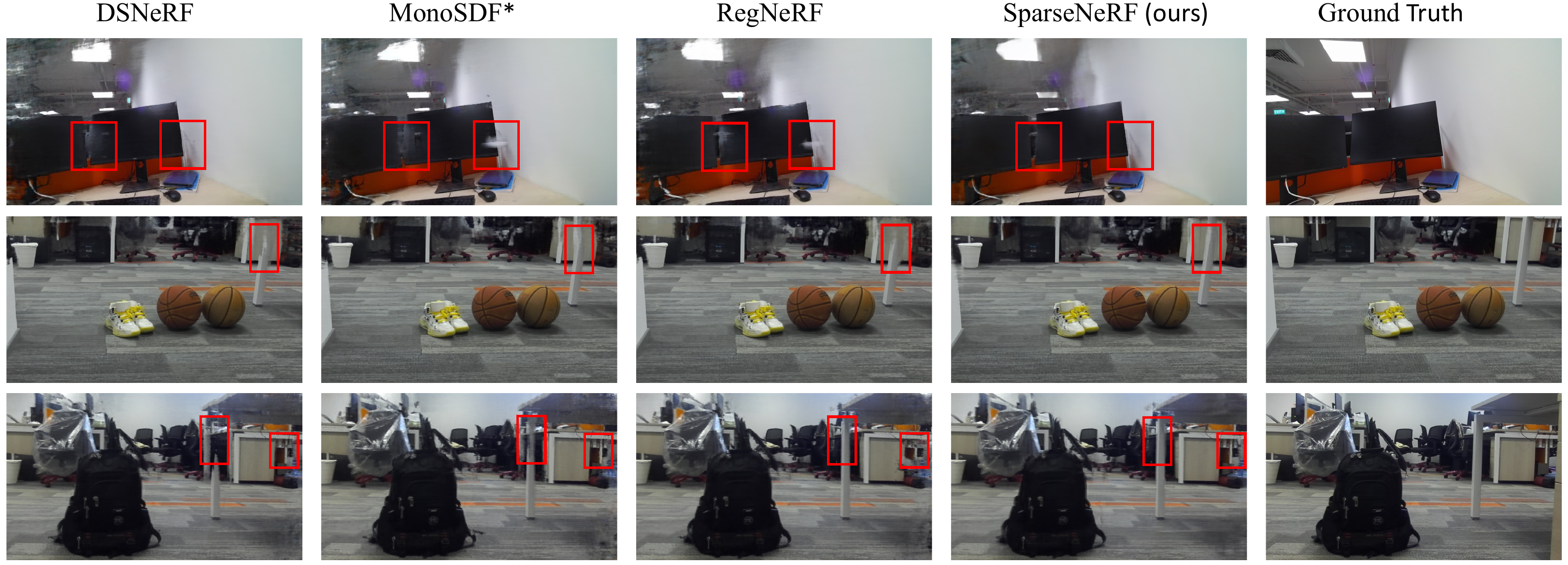}
  \caption{Visual comparisons on the NVS-RGBD dataset with three views. Red boxes denote compared regions. SparseNeRF achieves consistent improvement in different scenes.}
  \label{fig:NVS_comparison_supp}
\end{figure*}

\begin{figure*}[t]
  \centering
  \includegraphics[width=0.95\linewidth]{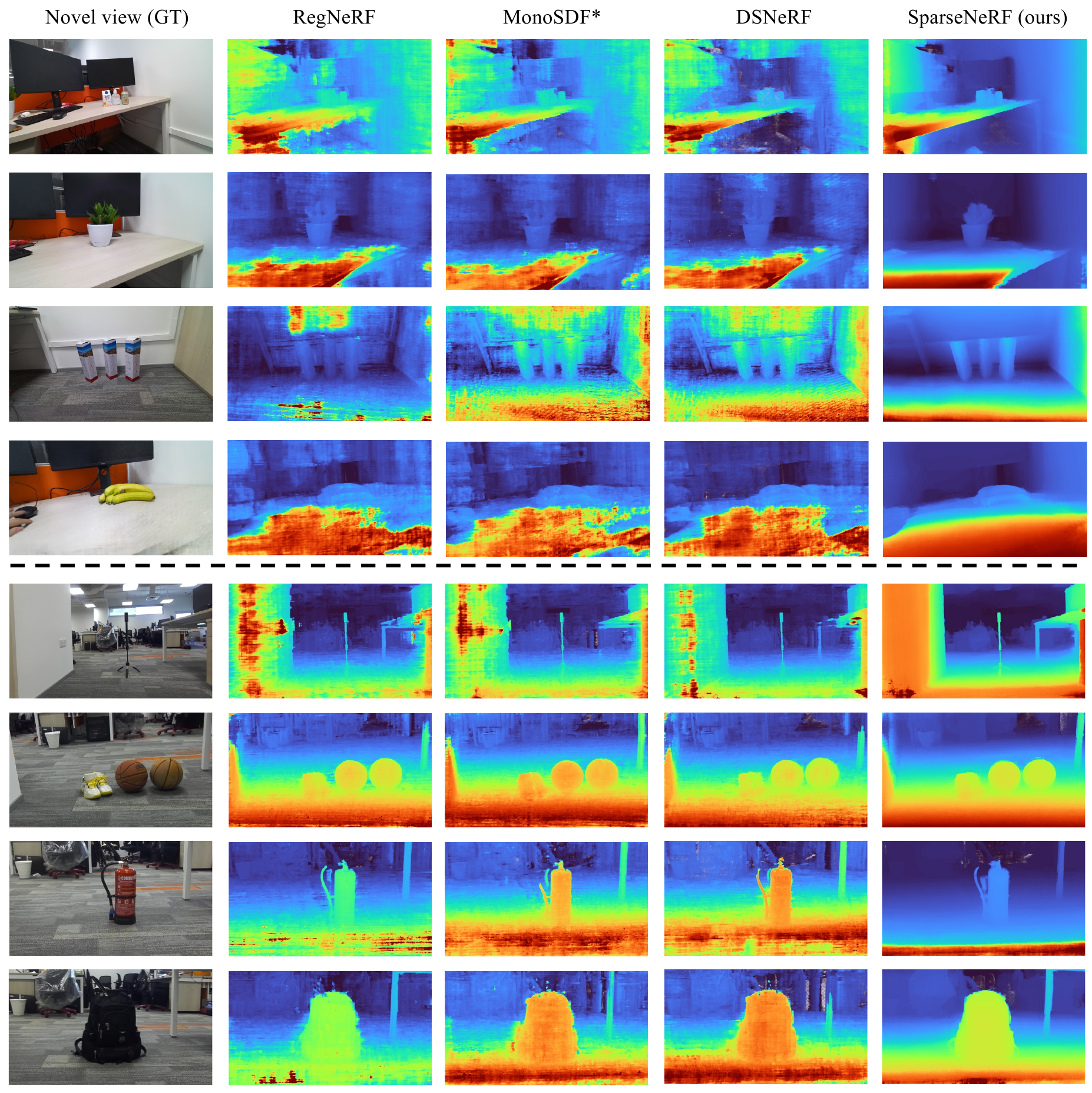}
  \vspace{-10pt}
  \caption{Visual comparisons of predicted geometry. Top: captured by Azure Kinect. Bottom: captured by ZED 2.}
  \label{fig:compare_geometry}
  \vspace{-10pt}
\end{figure*}

The examples of the NVS-RGBD dataset are shown in \textbf{Figure \ref{fig:nvs_examples}}. As for Azure Kinect, the edges of objects often contain lots of noise. Some regions are directly masked by sensors. As for ZED 2, the edges of the depth maps are inaccurate. In \textbf{Figure \ref{fig:coarse_depth}}, we observe that the depth maps of ZED 2 look smooth but unstable and incorrect when observing time jittering. We masked out the uncertain black regions of depth maps for training. Moreover, the scale of these depth maps is not linearly scaled to the ground-truth depth maps predicted by dense-view NeRFs. As for the depth maps of the iPhone, they are more smooth than Kinect and ZED 2. The drawback is that the distance between objects and the iPhone is up to 2.5 meters.

\section*{C. More Implementation Details}

In \textbf{Tables \ref{tab:Comparisons_llff}} and \textbf{\ref{tab:Comparisons_dtu}}, and \textbf{Figures} \textbf{\ref{fig:llff_comparison}} and \textbf{\ref{fig:dtu_comparison}}, we reproduce most of the prior arts as the supplementary Material of RegNeRF. we implement DSNeRF as the official codebase. In \textbf{Table \ref{fig:NVS_comparison}}, MonoSDF is mainly designed for surface reconstruction. We apply the depth consistency loss to RegNeRF for novel view synthesis of RGB images for comparison. The depth consistency loss is a scale-invariant loss constrained in local patches. In \textbf{Table \ref{tab:nvs_rgbd}}, we can see that MonoSDF obtains a lower depth error than RegNeRF, but fails to achieve better novel view synthesis of RGB images.

\section*{D. More Visual Comparisons}
In \textbf{Figures \ref{fig:llff_comparison}}, \textbf{\ref{fig:dtu_comparison}}, and \textbf{\ref{fig:NVS_comparison}} of the main body of the paper, we give two examples of visual comparisons. In this material, \textbf{Figures \ref{fig:llff_comparison_supp}}, \textbf{\ref{fig:dtu_comparison_supp}}, and \textbf{\ref{fig:NVS_comparison_supp}} show more visual examples on the LLFF, DTU, and NVS-RGBD datasets, respectively. In particular, we compare five representative methods on LLFF and DTU, respectively. Among them, MipNeRF is a state-of-the-art NeRF method designed for conventional dense-view NeRF, which is optimized by color reconstruction. Without extra geometric regularization, MipNeRF fails to synthesize good novel views. As shown in  \textbf{Figures \ref{fig:llff_comparison_supp}} and \textbf{\ref{fig:dtu_comparison_supp}}, it is observed that positions of objects predicted by MipNeRF might be shifted due to the under-constrained problem. PixelNeRF(ft) and MVSNeRF(ft) require pre-training on extra scenes and fine-tuning on a target scene. Thanks to the pre-training, PixelNeRF(ft) and MVSNeRF(ft) greatly improve few-shot NeRFs. Compared with PixelNeRF(ft) and MVSNeRF(ft), RegNeRF and SparseNeRF do not require extra scenes for pre-training. RegNeRF optimizes few-shot NeRF by adding a continuous geometric constraint from unseen viewpoints such that the depth values of neighbor pixels should be as close as possible. RegNeRF achieves a new state-of-the-art performance but fails to guarantee the correct geometry of objects. Unlike RegNeRF, the proposed SparseNeRF exploits coarse depth maps of pre-trained depth models in the experiments. Thanks to the distillation of depth ranking and spatial continuity of pe-trained depth models, SparseNeRF significantly improves few-shot NeRFs.

In \textbf{Figures \ref{fig:llff_comparison_supp}} and \textbf{\ref{fig:dtu_comparison_supp}}, SparseNeRF distills depth priors from pre-trained depth models. In \textbf{Figure \ref{fig:NVS_comparison_supp}}, we use depth sensors to collect coarse depth maps. We compare two depth-based methods and the best few-shot NeRF method RegNeRF. Among them, DSNeRF adopts sparse 3D points derived by the COLMAP algorithm \cite{schoenberger2016sfm}. We implement depth-supervised loss of DSNeRF upon RegNeRF. MonoSDF is originally proposed for surface reconstruction. We implement the depth consistency loss upon RegNeRF, which encourages the predicted depth to be scale-invariant to coarse depth maps. The drawback of this method is the strong assumption about the scale-invariant depth constraint. The cheap depth maps from pre-trained single-view depth models or consumer-level depth sensors are incorrect. They only provide coarse geometry information. Compared with the two depth-based NeRFs and the best few-shot NeRF RegNeRF, the proposed SparseNeRF achieves a better visual effect.

\section*{E. Visual Comparisons of Predicted Geometry}

In \textbf{Table \textbf{\ref{tab:nvs_rgbd}}} of the paper, we compare the scale-invariant depth errors of four few-shot NeRF methods. The results show that the SparseNeRF significantly outperforms RegNeRF, i.e., from 5.9$\times 10^{-3}$ to 3.9$\times 10^{-3}$ on the scenes captured by Kinect and from 3.5$\times 10^{-3}$ to 2.4$\times 10^{-3}$ on the scenes captured by ZED 2. DSNeRF and MonoSDF also improve RegNeRF on depth errors. The depth error indicates the quality of the reconstructed geometry.

In this material, we study the qualitative results of geometry, as shown in Figure \textbf{\ref{fig:compare_geometry}}. The top four rows show the comparisons on the RGBD data of Kinect. The bottom four rows are for ZED 2. Compared with RegNeRF, DSNeRF uses sparse 3D points for depth supervision. MonoSDF imposes a local scale-invariant loss computed by coarse depth maps and predicted depth of NeRFs. Compared with RegNeRF, MonoSDF, and DSNeRF, SparseNeRF yields much better geometry. The sub-optimal results of MonoSDF can be attributed to the fact that coarse depth maps are not able to be linearly scaled to ground-truth depth maps. In the next section, we will analyze the scale-invariant depth errors of coarse depth maps.

\end{document}